\documentclass{article}

\usepackage{nac_preprint}
\usepackage{float} 

\usepackage[T1]{fontenc}
\usepackage[utf8]{inputenc}
\usepackage{nicefrac}       
\usepackage{microtype}      

\usepackage{amsfonts}       
\usepackage{amsmath}
\usepackage{amssymb}
\usepackage{mathtools}
\usepackage[toc,page]{appendix}
\usepackage{enumitem}
\usepackage{graphics}
\usepackage{graphicx}
\usepackage{xcolor}
\usepackage{subcaption}

\usepackage[export]{adjustbox}
\usepackage{algorithm}
\usepackage{algorithmicx}
\usepackage[noend]{algpseudocode}
\algnewcommand{\LineComment}[1]{\State \(//\) #1}
\algnewcommand{\RLineComment}[1]{\State \(\triangleright\) #1}

\usepackage{etoolbox}
\usepackage{tikz}
\usetikzlibrary{tikzmark}
\usetikzlibrary{calc}

\errorcontextlines\maxdimen

\newcommand{\ALGtikzmarkcolor}{black}
\newcommand{\ALGtikzmarkextraindent}{4pt}
\newcommand{\ALGtikzmarkverticaloffsetstart}{-.5ex}
\newcommand{\ALGtikzmarkverticaloffsetend}{-.5ex}
\makeatletter
\newcounter{ALG@tikzmark@tempcnta}

\newcommand\ALG@tikzmark@start{%
    \global\let\ALG@tikzmark@last\ALG@tikzmark@starttext%
    \expandafter\edef\csname ALG@tikzmark@\theALG@nested\endcsname{\theALG@tikzmark@tempcnta}%
    \tikzmark{ALG@tikzmark@start@\csname ALG@tikzmark@\theALG@nested\endcsname}%
    \addtocounter{ALG@tikzmark@tempcnta}{1}%
}

\def\ALG@tikzmark@starttext{start}
\newcommand\ALG@tikzmark@end{%
    \ifx\ALG@tikzmark@last\ALG@tikzmark@starttext
    \else
        \tikzmark{ALG@tikzmark@end@\csname ALG@tikzmark@\theALG@nested\endcsname}%
        \tikz[overlay,remember picture] \draw[\ALGtikzmarkcolor] let \p{S}=($(pic cs:ALG@tikzmark@start@\csname ALG@tikzmark@\theALG@nested\endcsname)+(\ALGtikzmarkextraindent,\ALGtikzmarkverticaloffsetstart)$), \p{E}=($(pic cs:ALG@tikzmark@end@\csname ALG@tikzmark@\theALG@nested\endcsname)+(\ALGtikzmarkextraindent,\ALGtikzmarkverticaloffsetend)$) in (\x{S},\y{S})--(\x{S},\y{E});%
    \fi
    \gdef\ALG@tikzmark@last{end}%
}

\apptocmd{\ALG@beginblock}{\ALG@tikzmark@start}{}{\errmessage{failed to patch}}
\pretocmd{\ALG@endblock}{\ALG@tikzmark@end}{}{\errmessage{failed to patch}}
\makeatother

\title{The Predictive Forward-Forward Algorithm} 

\author{%
  Alexander Ororbia \\
  Rochester Institute of Technology \\
  \texttt{ago@cs.rit.edu}
  \And
  Ankur Mali\\
  University of South Florida\\
  \texttt{ankurarjunmali@usf.edu}
}

\begin{document}

\maketitle


\begin{abstract}
We propose the predictive forward-forward (PFF) algorithm for conducting credit assignment in neural systems. Specifically, we design a novel, dynamic recurrent neural system that learns a directed generative circuit jointly and simultaneously with a representation circuit. Notably, the system integrates learnable lateral competition, noise injection, and elements of predictive coding, an emerging and viable neurobiological process theory of cortical function, with the forward-forward (FF) adaptation scheme. Furthermore, PFF efficiently learns to propagate learning signals and updates synapses with forward passes only, eliminating key structural and computational constraints imposed by backpropagation-based schemes. Besides computational advantages, the PFF process could prove useful for understanding the learning mechanisms behind biological neurons that use local signals despite missing feedback connections. We run experiments on image data and demonstrate that the PFF procedure works as well as backpropagation, offering a promising brain-inspired  algorithm for classifying, reconstructing, and synthesizing data patterns.

\small{}
\end{abstract}

\section{Introduction}
\label{sec:intro}

The algorithm known as backpropagation of errors \cite{bp2, linnainmaa1970representation}, or ``backprop'' for short, has long faced criticism concerning its neurobiological plausibility \cite{crick1989recent, gardner1993neurobiology, shepherd1990significance, marblestone2016toward, grossberg1987competitive}. Despite powering the tremendous progress and success behind deep learning and its every-growing myriad of promising applications \cite{silver2016mastering,floridi2020gpt}, it is improbable that backprop is a viable model of learning in the brain, such as in cortical regions. Notably, there are both practical and biophysical issues \cite{grossberg1987competitive, marblestone2016toward}, and, among these issues, there is a lack of evidence that:
\textbf{1)} neural activities are explicitly stored to be used later for synaptic adjustment, 
\textbf{2)} error derivatives are backpropagated along a global feedback pathway to generate teaching signals, 
\textbf{3)} the error signals move back along the same neural pathways used to forward propagate information, and, 
\textbf{4)} inference and learning are locked to be largely sequential (instead of massively parallel).
Furthermore, when processing temporal data, it is certainly not the case that the neural circuitry of the brain is unfolded backward through time to adjust synapses \cite{ororbia2018continual} (as in backprop through time).

\begin{figure*}[!t] 
 \centering
\includegraphics[width=0.9\textwidth]{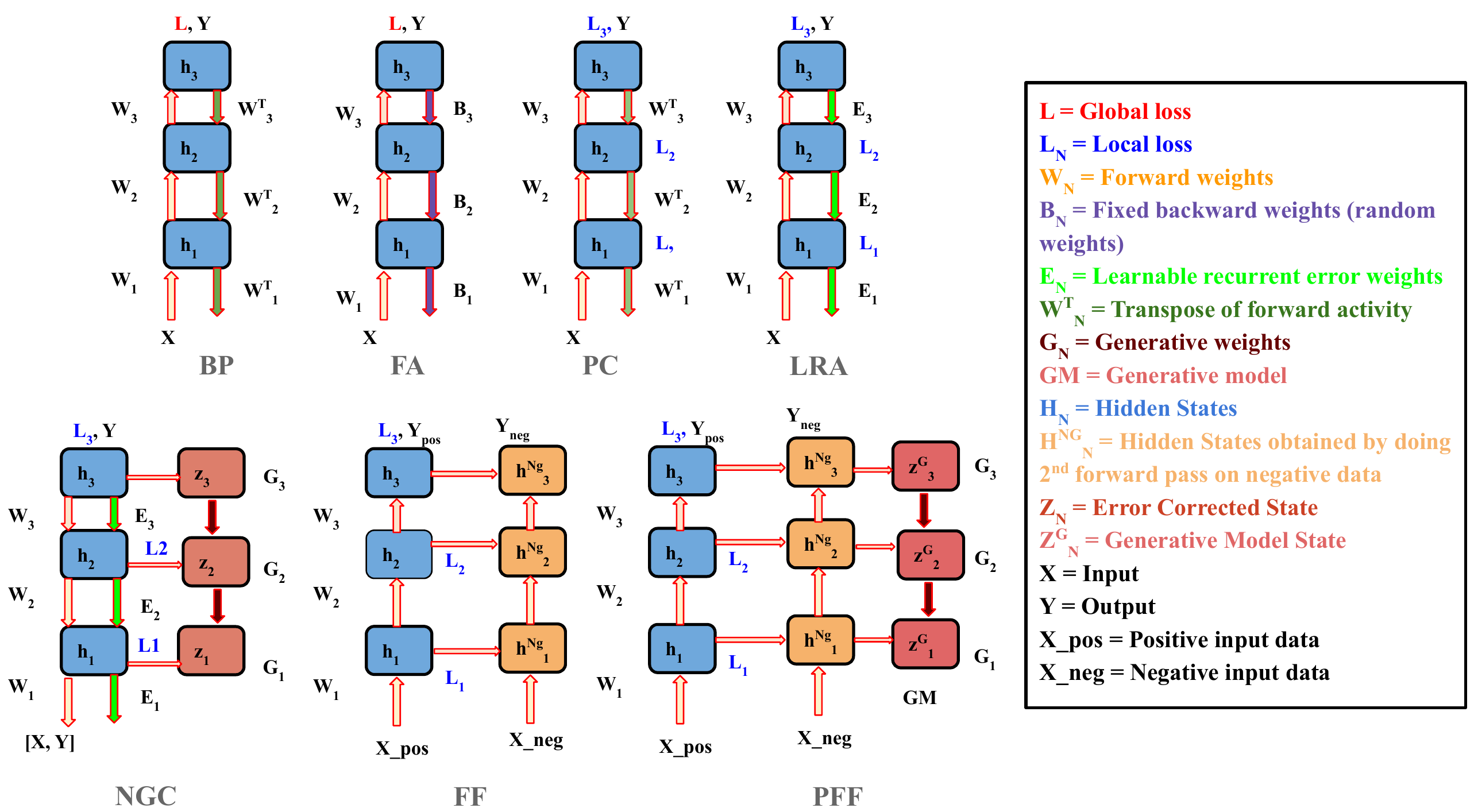}
\caption{Comparison of credit assignment algorithms that relax constraints imposed by backpropagation of errors (BP). Algorithms visually depicted include feedback alignment (FA) \cite{lillicrap2014random}, predictive coding (PC) \cite{rao1999predictive,salvatori2022reverse}, local representation alignment (LRA) \cite{ororbia2018biologically}, neural generative coding (NGC) \cite{ororbia&mali2019lifelong,ororbia2022neural}, the forward-forward procedure (FF) \cite{hinton2022forward}, and the predictive forward-forward algorithm (PFF).}
\label{fig:learningalgo_comp}
\end{figure*}

Recently, there has been a growing interest in the research domain of brain-inspired computing, which focuses on developing algorithms and computational models that attempt to circumvent or resolve critical issues such as those highlighted above. Among the most powerful and promising ones is predictive coding (PC) \cite{helmholtz1924treatise, rao1999predictive,friston2010free,bastos2012canonical,salvatori2021associative,ororbia2022neural}, and among the most recent ones is the forward-forward (FF) algorithm \cite{hinton2022forward}. These alternatives offer different means of conducting credit assignments with performance similar to backprop, but to the contrary, are more likely consistent with and similar to real biological neuron learning (see Figure \ref{fig:learningalgo_comp} for a graphical depiction and comparison of respective credit assignment setups). 
This paper will propose a novel model and learning process, the \textbf{predictive forward-forward (PFF)} process, that generalizes and combines FF and PC into a robust stochastic neural system that simultaneously learns a representation and generative model in a biologically-plausible fashion. Like the FF algorithm, the PFF procedure offers a promising, potentially helpful model of biological neural circuits, a potential candidate system for low-power analog hardware and neuromorphic circuits, and a potential backprop-alternative worthy of future investigation and study.\footnote{Code for the PFF algorithm can be found at: \texttt{https://github.com/ago109/predictive-forward-forward}} 

\section{Predictive Forward-Forward Learning}
\label{sec:method}

The brain-inspired credit assignment process that we will design and study is called the predictive forward-forward (PFF) algorithm, which is a generalization of the FF algorithm \cite{hinton2022forward}. At a high level, the PFF process consists of two neural structures or circuits, i.e., a representation circuit (parameterized by $\Theta_r$) that focuses on acquiring distributed representations of data samples and a top-down generative circuit (parameterized by $\Theta_g$) that focuses on learning how to synthesize data given the activity values of the representation circuit. Thus, the PFF process can be characterized as a complementary system that jointly learns a classifier and generative model. We will first define the notation used in this paper, then proceed to describe the inference and learning mechanics of the representation and generative circuits.

\noindent 
\textbf{Notation:} 
We use $\odot$ to indicate a Hadamard product, $\cdot$ to denote a matrix/vector multiplication. $(\mathbf{v})^T$ is the transpose of $\mathbf{v}$. Matrices/vectors are depicted in bold font, e.g., matrix $\mathbf{M}$ or vector $\mathbf{v}$ (scalars shown in italics). $\mathbf{z}_j$ will refer to extracting $j$th scalar from vector $\mathbf{z}$. $<\mathbf{a},\mathbf{b}>$ denotes vector concatenation along the column dimension. 
Finally, $||\mathbf{v}||_2$ denotes the Euclidean norm of vector $\mathbf{v}$. Sensory input has shape $\mathbf{x} \in \mathcal{R}^{J_0 \times 1}$ ($J_0$ is the number of input features), the label has shape $\mathbf{y} \in \mathcal{R}^{C \times 1}$ ($C$ is the number of classes), and a neural layer has shape $\mathbf{z}^\ell \in \mathcal{R}^{J_\ell \times 1}$ ($J_\ell$ is the number of neurons for $\ell$). 

\subsection{The Forward-Forward Learning Rule}
\label{sec:ff_rule}

The PFF process, much like the FF algorithm when it is applied to a recurrent network, involves adjusting the synaptic efficacies of a group of neurons by measuring their ``goodness'', i.e.,  the probability that their activity indicates that an incoming signal comes from the target training data distribution (or the ``positive class''). Formally, for layer $\ell$ in an $L$-layered neural system, we calculate the goodness as the sum of the squared activities for a given neural activity vector $\mathbf{z}^\ell$ and compare it to threshold value $\theta_z$ in one of two ways:
\begin{align}
    p(c=1)_\ell &= \frac{1}{1 + \exp \big( -(\sum^{J_\ell}_{j=1} (\mathbf{z}^\ell_j)^2 - \theta_z ) \big)}, \; \mbox{or, } \label{eqn:goodness_max} \\ 
    p(c=1)_\ell &= \frac{1}{1 + \exp \big( -(\theta_z - \sum^{J_\ell}_{j=1} (\mathbf{z}^\ell_j)^2 ) \big)} \label{eqn:goodness_min}
\end{align}
where $p(c=1)_\ell$ indicates the probability that the data comes from the data distribution (i.e., positive data, where the positive class is labeled $c = 1$) while the probability that the data does not come from the training data distribution is $p(c=0)_\ell = 1 - p(c=1)_\ell$. Note that $p(c)_\ell$ indicates the probability assigned to a layer $\ell$ of neurons in a system/network.  
This means the cost function that any layer is trying to solve/optimize is akin to a binary class logistic regression problem formulated as:
\begin{align}
    \mathcal{L}(\Theta^\ell) = -\frac{1}{N}\sum^N_{i=1} c_i \log p(c_i=1)_\ell + (1 - c_i) \log p(c_i=0)_\ell \label{eqn:rep_loss} 
\end{align}
where the binary label $c_i$ (the label for the $i$th datapoint $\mathbf{x}_i$) can be generated correctly and automatically if one formulates a generative process for producing negative data samples. Data patterns sampled from the training set $\mathbf{x}_j \sim \mathcal{D}_{train}$ can be labeled as $c_j = 1$ and patterns sampled outside of $\mathcal{D}_{train}$ (from the negative data generating process) can be labeled as $c_j = 0$. Crucial to the success of the FF procedure is the design of a useful negative data distribution, much as is done for noise contrastive estimation \cite{gutmann2010noise}. 

It is important to notice that the FF learning rule is local -- this means that the synapses of any particular layer of neurons can be adjusted independently of others. The rule's form is different from a classical Hebbian update \cite{hebb1949organization} (which produces a weight change through a product of incoming and outgoing neural activities), given that this synaptic adjustment requires knowledge across a group of neurons (goodness depends on the sum of squares of the activities of a group rather than an individual unit) and integrates contrastive learning into the dynamics. 
Synaptic updates are calculated by taking the gradient of Equation \ref{eqn:rep_loss}, i.e., $\frac{\partial \mathcal{L}(\Theta^\ell)}{\partial \Theta^\ell}$. 
In effect, a neural layer optimizes Equation \ref{eqn:rep_loss} by either maximizing the squared activities of a layer (to be above $\theta_z$, i.e., Equation \ref{eqn:goodness_max}) or minimizing the squared activities (to be below $\theta_z$, i.e., Equation \ref{eqn:goodness_min}).

\subsection{The Representation Circuit}
\label{sec:ff_rnn}

In order to take advantage of the FF learning rule and to model contextual prediction, we propose a neural circuit structured as a recurrent neural network (RNN) similar in spirit to the one in \cite{hinton2022forward}, where, at each layer, top-down and bottom-up influences are combined to compute layer $\ell$'s activity, much akin to the inference process of a deep Boltzmann machine \cite{salakhutdinov2010efficient}. The core parameters of this representation circuit are housed in the construct $\Theta_r = \{\mathbf{W}^1,\mathbf{W}^2,...,\mathbf{W}^L\}$ (referred to as representation parameters). Note that no extra classification-specific parameters are included in our model (in contrast to the model of \cite{hinton2022forward}), although incorporating these is straightforward.\footnote{
If classification-specific parameters are desired, one could include an additional set of synaptic weights $\Theta^d = \{\mathbf{W},\mathbf{b}\}$ that take in as input the top-most (normalized) activity $\text{LN}(\mathbf{z}^L)$ of the recurrent representation circuit to make a rough prediction of the label distribution over $\mathbf{y}$, i.e, $p(y=i|\text{LN}(\mathbf{z}^L)) = \exp(\mathbf{W} \cdot \text{LN}(\mathbf{z}^L) + \mathbf{b})_i/\big( \sum_c \exp(\mathbf{W} \cdot \text{LN}(\mathbf{z}^L) + \mathbf{b})_c \big)$. This would make the recurrent model of this work much more similar to that of \cite{hinton2022forward}. 
Softmax parameters $\mathbf{W}$ and $\mathbf{b}$ would then be adjusted by taking the relevant gradients of the objective $\mathcal{L}^y(\mathbf{W},\mathbf{b}) = -\log p(y=i | \text{LN}(\mathbf{z}^L))$.
} 

To compute any layer's activity within this representation circuit, top-down, lateral, and bottom-up messages are combined with an interpolation of the layer's activity at the previous time step. Specifically, in PFF, this is done in the following manner:
\begin{align}
    \mathbf{z}^\ell(t) =  \beta \Big( \phi^\ell \big( &\mathbf{W}^\ell \cdot \text{LN}( \mathbf{z}^{\ell-1}(t-1) ) + \mathbf{V}^\ell \cdot \text{LN}( \mathbf{z}^{\ell+1}(t-1) ) \nonumber \\
    &- \mathbf{L}^\ell \cdot \text{LN}(\mathbf{z}^\ell(t-1)) + \mathbf{\epsilon}^\ell_r \Big) + (1 - \beta) \mathbf{z}^\ell(t-1) \label{eqn:ff_state_update}
\end{align}
where $\epsilon^\ell_r \sim \mathcal{N}(0,\sigma)$ is injected, centered Gaussian noise and $\mathbf{z}^0(t-1) = \mathbf{x}$. 
Notably, our circuit's dynamics directly integrate learnable lateral/cross inhibition and self-excitation through the matrix $\mathbf{L}^\ell$, which is further factorized as follows:
\begin{align}
    \mathbf{L}^\ell = \text{ReLU}( \mathbf{\hat{L}}^\ell ) \odot \mathbf{M}^\ell \odot (1 - \mathbf{I}) + \text{ReLU}( \mathbf{\hat{L}}^\ell ) \odot \mathbf{I}
\end{align}
where $\mathbf{\hat{L}}^\ell \in \mathcal{R}^{J_\ell \times J_\ell}$ is a learnable parameter matrix, $\mathbf{M}^\ell \in \{0,1\}^{J_\ell \times J_\ell}$ is a binary masking matrix that enforces a particular lateral neural competition pattern (see the Appendix for details), and $\mathbf{I}^\ell \in \{0,1\}^{J_\ell \times J_\ell}$ is an identity matrix. 
We set the activation function $\phi^\ell()$ for each layer $\ell$ to be the linear rectifier, i.e.,  $\phi^\ell(\mathbf{v}) = \text{max}(0, \mathbf{v})$. The interpolation coefficient $\beta$ controls the integration of the state $\mathbf{z}^{\ell}$ over time (i.e., the new activity state at time $t$ is a convex combination of the newly proposed state and the previous value of the state at $t-1$). Notice that this interpolation is similar to the ``regression'' factor introduced into the recirculation algorithm \cite{hinton1988learning}, a classical local learning process that made use of carefully crafted autoencoders to generate the signals needed for computing synaptic changes.   
$\text{LN}(\mathbf{z})$ is a layer normalization function applied to the activity vector $\mathbf{z}^\ell$, i.e., $\text{LN}(\mathbf{z}^\ell) = \mathbf{z}^\ell/(||\mathbf{z}^\ell||_2 + \epsilon)$ ($\epsilon$ is a small numerical stability factor for preventing division by zero). 
Note that the topmost layer of the representation circuit is clamped to a context vector $\mathbf{y}$ (which could be the output of another circuit or clamped to a data point's label), i.e., $\mathbf{z}^{L+1} = \mathbf{y}$\footnote{It is important to scale the label/context vector by a factor of about $5$, i.e., the topmost layer activity would be $\mathbf{z}^{L+1} = \mathbf{y} * 5$ (Geoffrey Hinton, personal communication, Dec 12, 2022).}, while the bottom layer is clamped to sensory input, i.e., $\mathbf{z}^0(t) = \mathbf{x}(t)$ ($\mathbf{x}(t)$ could be a video frame or a repeated copy of a static image $\mathbf{x}$). 
Equation \ref{eqn:ff_state_update} depicts a synchronous update of all of the layer-wise activities, but the RNN could also be implemented asynchronously \cite{hinton2022forward}, i.e., first update all even-numbered layers given the activities of the odd-numbered layers followed by updating the values of the odd-numbered layers given the new values of the even-numbered layers, as was done in the generative stochastic networks of \cite{bengio2014deep}.

To create the negative data needed to train this system, we disregard the current class indicated by the label $\mathbf{y}$ of the positive data $\mathbf{x}^p$ and create an incorrect ``negative label'' $\mathbf{y}^n$ by randomly (uniformly) sampling an incorrect class index, excluding the correct one.\footnote{This deviates from negative label creation in the FF algorithm \cite{hinton2022forward}, which chose an incorrect class index in proportion to the probabilities produced by a forward pass of the classification-specific weights. This was not needed for the PFF process.} 
The final input to be presented to the representation circuit is created by concatenating the positive and negative sample, i.e., $\mathbf{x} = <\mathbf{x}, \mathbf{x}>$ (notice that positive image pixels are reused as negative data when labels are available), $\mathbf{y} = <\mathbf{y}, \mathbf{y}^n>$, and $\mathbf{c} = <1,0>$. 

Upon creating $(\mathbf{x},\mathbf{y},\mathbf{c})$, Equation \ref{eqn:ff_state_update} is run several times ($T = 8$ to $10$), 
akin to the stimulus processing window simulated in  predictive coding systems \cite{rao1999predictive,ororbia2022neural}. Each time Equation \ref{eqn:ff_state_update} is run, the updates for the synapses for layer $\ell$ are calculated according to the following local update rules:
\begin{align}
    \Delta \mathbf{W}^\ell &= \Big( 2 \frac{\partial \mathcal{L}(\Theta^\ell)}{\partial \sum^{J_\ell}_j (\mathbf{z}^\ell_j)^2} \odot \mathbf{z}^\ell \Big) \cdot \big( \text{LN}(\mathbf{z}^{\ell-1}) \big)^T, \; \mbox{and, } \\ 
    \Delta \mathbf{V}^\ell &= \Big( 2 \frac{\partial \mathcal{L}(\Theta^\ell)}{\partial \sum^{J_\ell}_j (\mathbf{z}^\ell_j)^2} \odot \mathbf{z}^\ell \Big) \cdot \big( \text{LN}(\mathbf{z}^{\ell+1}) \big)^T \\ 
    \Delta \mathbf{\hat{L}}^\ell &= \Big( \Big( 2 \frac{\partial \mathcal{L}(\Theta^\ell_r)}{\partial \sum^{J_\ell}_j (\mathbf{z}^\ell_j)^2} \odot \mathbf{z}^\ell \Big) \cdot \big( \text{LN}(\mathbf{z}^{\ell}) \big)^T \Big) \odot \frac{\partial \mathbf{L}^\ell}{\partial \mathbf{\hat{L}}^\ell}
\end{align}
which can then be applied to adjust the relevant parameters, i.e., $\mathbf{W}^\ell$ and $\mathbf{V}^\ell$, via methods such as stochastic gradient descent (SGD) with momentum or Adam \cite{kingma2014adam}. In principle, the neural layers of the representation circuit are globally optimizing the objective $\mathcal{L}(\Theta_r) = \sum^L_{\ell=1} \mathcal{L}(\Theta^\ell = \mathbf{W}^\ell)$ (the summation of local goodness functions).

\noindent
\textbf{On Classifying Sensory Patterns:} One might observe that our representation circuit does not include discriminatory parameters to classify inputs directly. Nevertheless, given that the supervised target $\mathbf{y}$ (for positive data samples) is used as context to mediate the top-most latent representations of the recurrent circuit above, the representation system should acquire distributed representations that implicitly encode label information. 
To take advantage of the discriminative information encoded in PFF's representations, as was also done in the FF algorithm, we may classify by executing an inference process similar to that of hybrid Boltzmann machine models \cite{larochelle2008classification,ororbia2015online}. 

In order to classify an input $\mathbf{x}$, we iterate over all possible (one-hot) values that $\mathbf{y}$ could be, starting with the first class index. Specifically, for any chosen $\mathbf{y}$ (such as the one-hot encoding of class index $i$), we run Equation \ref{eqn:ff_state_update} for the representation circuit for $T$ steps and then record the global goodness across the layers in the middle three iterations (from $T/2-1$ to $T/2+1$), i.e., $\mathcal{G}_{y=i} = \frac{1}{3}\sum^{T/2+1}_{T/2-1} \frac{1}{L}\sum^L_{\ell=1} \theta_z - \sum^{J_\ell}_{j} (\mathbf{z^\ell_j}^2)$. This goodness calculation is made for all class indices $y = 1, 2,...,C$, resulting in $\{ \mathcal{G}_{y=1}, \mathcal{G}_{y=2}, ..., \mathcal{G}_{y=C} \}$ over which the argmax is applied to obtain the index of the class with the highest average goodness value. Note that, as mentioned in \cite{hinton2022forward}, if classification-specific parameters are included in PFF's representation circuit, then a feedforward pass could be used to obtain initial class probabilities. This pass could simplify the classification process by only requiring  goodness $\mathcal{G}_{y=i}$ to be computed for the top $M$ highest probabilities (side-stepping an expensive search over a massive number of classes). To estimate the label probability distribution under the representation circuit, as we do in this work, we run the goodness (logits) through the softmax, i.e., $p(y=i|\mathbf{x}) \sim \exp(\mathcal{G}_i)/(\sum_c \exp(\mathcal{G}_c))$.

\subsection{The Generative Circuit}
\label{sec:gen_circuit}

As mentioned before, the PFF algorithm incorporates the joint adaptation of a top-down directed generative model. This aspect of the PFF process is motivated by the generative nature of predictive processing (PP) models \cite{rao1999predictive,friston2010free}, particularly those that focus on learning a top-down generative model as in the framework of neural generative coding \cite{ororbia2022neural}. Crucially, we remark that jointly learning (in a biologically-plausible fashion) a generative feedback system could favorably provide a means of inspecting the content of the representations acquired by an FF-centric process and providing a plausible, alternative means for (internally) synthesizing negative data.

\begin{figure}[!t] 
 \centering
 \includegraphics[width=0.35\textwidth]{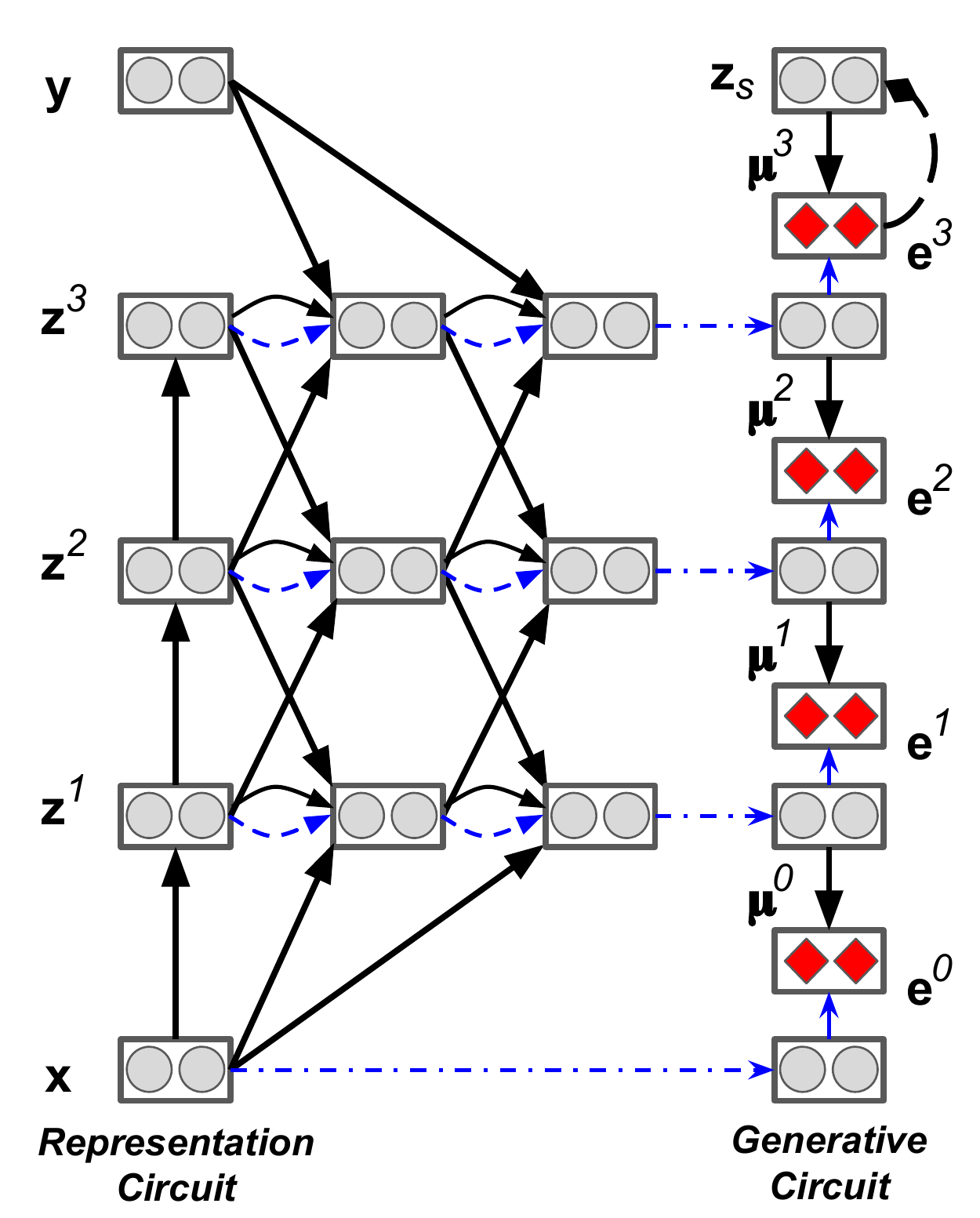}\\
 \includegraphics[width=0.3\textwidth]{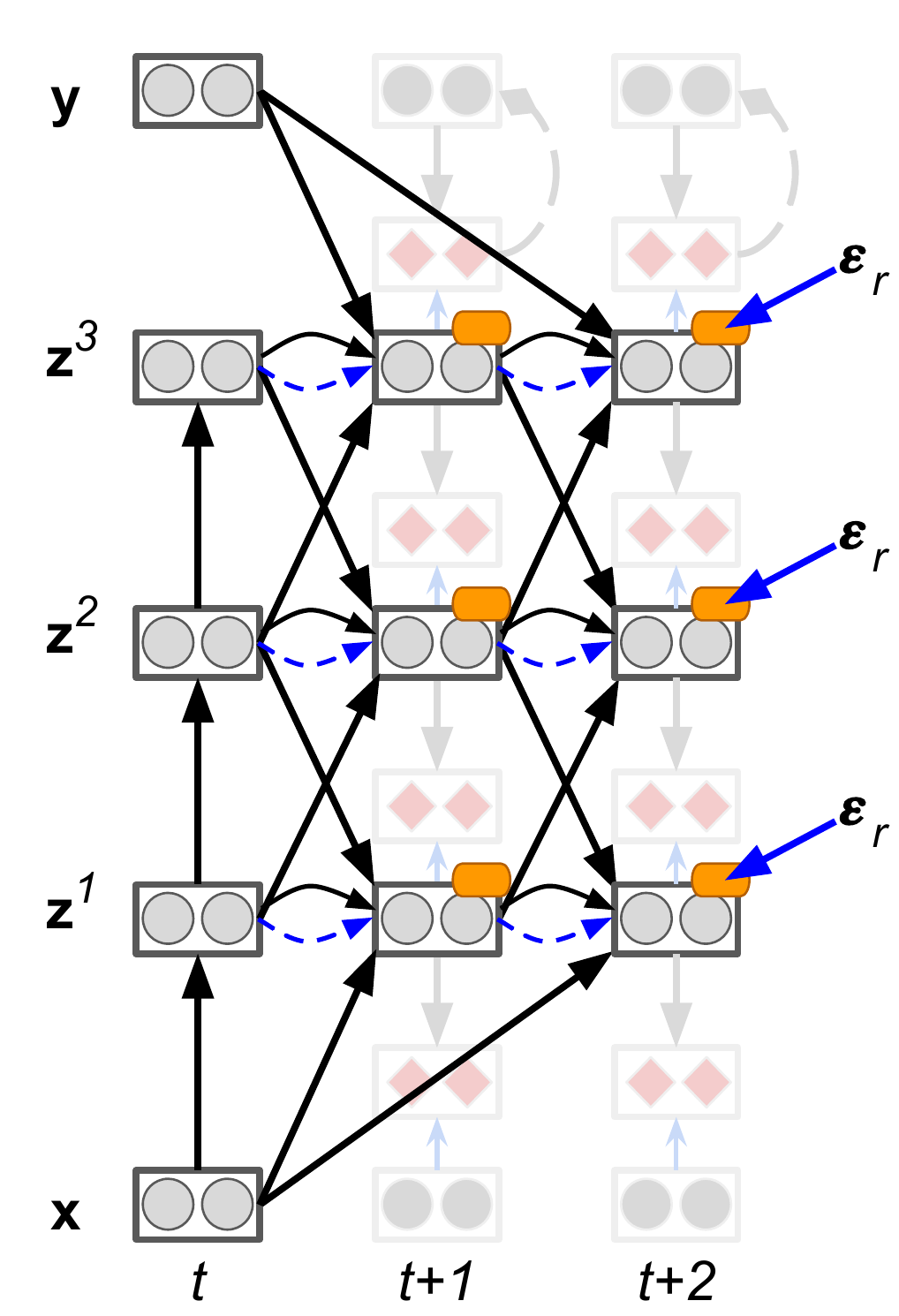}
 \hspace{0.55cm}
 \includegraphics[width=0.315\textwidth]{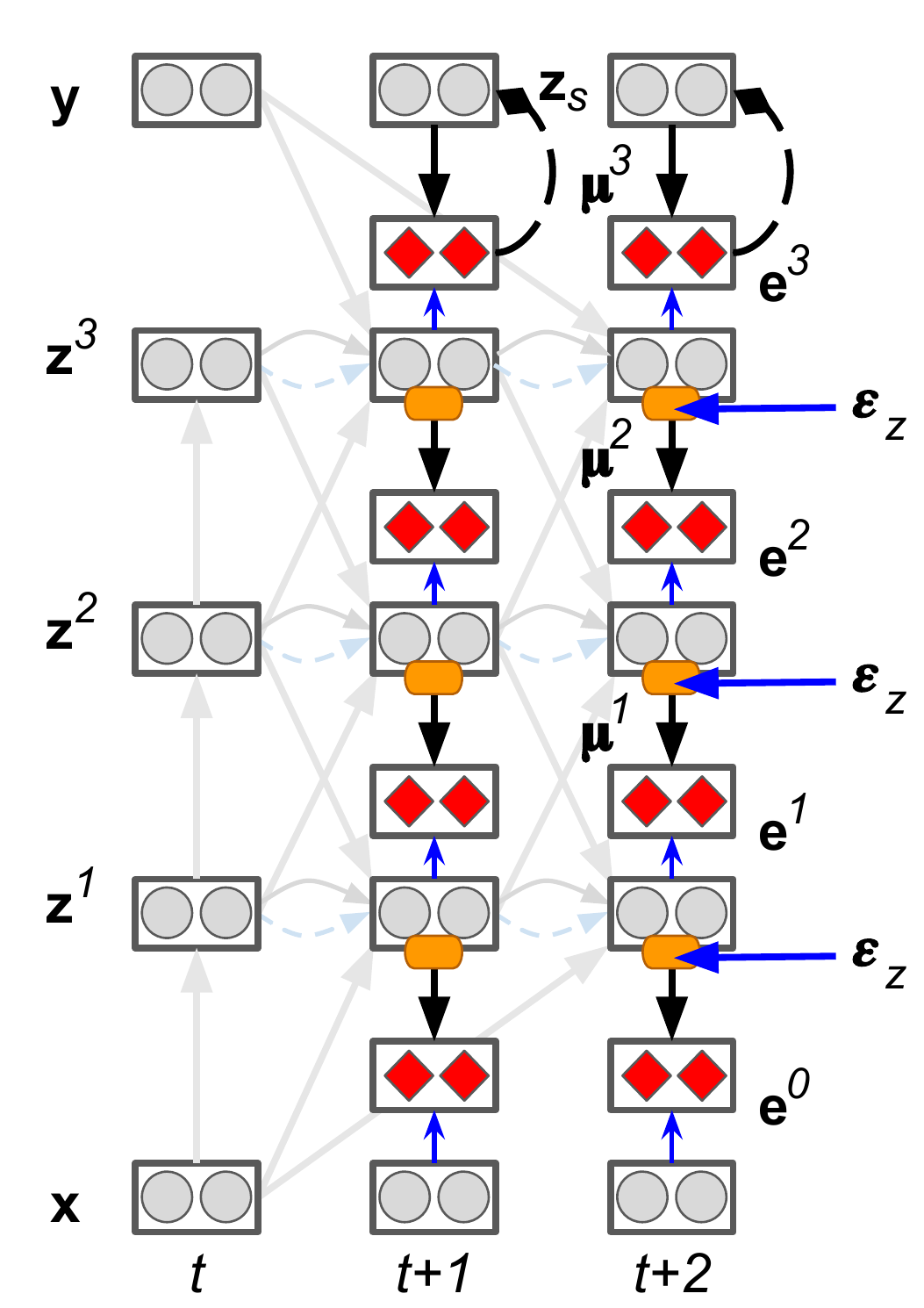}
 \vspace{-0.15cm}
\caption{(Top) The PFF algorithmic process depicted over three-time steps for a three hidden layer representation circuit coupled to a four-layer generative circuit ($\mathbf{z}_s$ is its latent variable). Solid arrows represent synapses 
and dash-dotted arrows depict state carry-over (direct copying). The dashed diamond curve depicts a feedback pathway, gray circles represent neural units, and red diamonds represent error neurons. Note: Since all system elements are adjusted dynamically, the generative circuit is run/updated each time alongside the representation circuit. 
(Bottom Left) The representation circuit is shown iteratively forming latent representations ($\mathbf{z}^\ell$) across time. (Bottom Right) The generative circuit is shown iteratively making local predictions ($\mathbf{\mu}^\ell$) of the representation circuit's layer-wise activities. Note that both of these processes happen simulatenously and noise is injected into each layer at each time step (blue arrow indicates injection of representation noise $\mathbf{\epsilon}^\ell_r$ or generative noise $\mathbf{\epsilon}^\ell_z$).}
\vspace{-0.5cm}
\label{fig:pff_process}
\end{figure}

The generative circuit, which is comprised of the set of synaptic parameters $\Theta_g = \{\mathbf{G}^1, \mathbf{G}^2,...,\mathbf{G}^L,\mathbf{G}^{L+1}\}$, attempts to learn how to predict, at each layer, a local region of neural activity, which, as we will see by design, facilitates simple error Hebbian updates (much like those calculated in a PP system). Formally, the objective that this generative circuit will attempt to optimize (for a single data point) is:
\begin{align}
    \mathcal{L}_g(\Theta_g) = \sum^{L}_{\ell=0} \mathcal{L}^{\ell}_g(\mathbf{G}^{\ell+1}) =  \sum^L_{\ell=0} \sum^{J_\ell}_{j=1} (\mathbf{\bar{z}}^\ell_j - \mathbf{z}^\ell_j(t))^2
\end{align}
where $\mathbf{z}^0 = \mathbf{x}$ (the bottom layer target is clamped to the data point being processed). 
Each layer of the generative circuit conducts the following computation:
\begin{align}
    \mathbf{\bar{z}}^\ell& = g^\ell(\mathbf{G}^{\ell+1} \cdot \text{LN}(\mathbf{\widehat{z}}^{\ell+1}) ), \\ 
    &\mbox{where, } \mathbf{\widehat{z}}^{\ell+1} = \phi^{\ell+1}(\mathbf{z}^{\ell+1}(t) + \epsilon^{\ell+1}_z) \; \mbox{and, } \; \mathbf{e}^\ell = \mathbf{\bar{z}}^\ell - \mathbf{z}^\ell(t) \\ 
    \mathbf{\bar{z}}^L &= g^L(\mathbf{G}^{L+1} \cdot \text{LN}(\mathbf{z}_s) ), \quad \text{// Topmost  layer  $\mathbf{z}_s$} \\ 
    &\mbox{where, } \; \mathbf{z}_s \leftarrow \mathbf{z}_s - \gamma \frac{\partial \mathcal{L}^L_g(\mathbf{G}^{L+1})}{\partial \mathbf{z}_s} 
\end{align}
where $\epsilon^\ell_z \sim \mathcal{N}(0,\sigma_z)$ is controlled (additive) activity noise injected into layer $\ell$ (with a small scale, i.e., $\sigma_z = 0.025$). $g^\ell()$ is the elementwise activation function applied to a generative layer's prediction, and in this work, we set this function for layers $\ell >= 1$ to be the linear rectifier except for the bottom one, which is set to be the clipped identity, i.e., $g^0(\mathbf{v}) = \text{HardClip}(\mathbf{v}, 0, 1)$. 
At each step of inference described in the section ``The Representation Circuit'', 
the synapses of the generative model (at each layer) are adjusted via the following Hebbian rule:
\begin{align}
    \Delta \mathbf{G}^\ell = \mathbf{e}^{\ell-1} \cdot \big( \text{LN}(\mathbf{z}^{\ell}(t)) \big)^T, \; \mbox{and, } \; \Delta \mathbf{G}^{L+1} = \mathbf{e}^L \cdot \big( \text{LN}(\mathbf{z}_s) \big)^T \mbox{.}
\end{align}
Notice that the topmost layer of the generative circuit (i.e., layer $L+1$) is treated a bit differently from the rest, i.e., the highest layer $\mathbf{z}_s$ predicts the topmost neural activity of the representation circuit $\mathbf{z}^L$ and is then adjusted by an iterative inference feedback scheme, much akin to that of sparse/predictive coding \cite{olshausen1997sparse,rao1999predictive,ororbia2022neural}. Once trained, synthesizing data from the generative circuit is done via ancestral sampling:
\begin{align}
    \mathbf{\bar{z}}^{L+1} &= \mathbf{z}_s \sim P(\mathbf{z}_s), \; \mbox{and,} \;
    \mathbf{\bar{z}}^\ell = g^\ell(\mathbf{G}^{\ell+1} \cdot \text{LN}(\mathbf{\bar{z}}^{\ell+1})), \; \ell = L,(L-1),...,0
\end{align}
where we choose the prior $P(\mathbf{z}_s)$ to be a Gaussian mixture model (GMM) with $10$ components, which, in this study, was retro-fit to samples of the trained system's topmost activities (acquired by running the training dataset $\mathcal{D}_{train}$ through the model), as was done for the top-down directed generative PP models of \cite{ororbia2022neural}. 
Note that for all circuits in PFF (both the representation and generative circuits), we treat the derivative of the linear rectifier activation function as a vector of ones with the same shape as the layer activity $\mathbf{z}^\ell$ (as was done in \cite{hinton2022forward}). 
The two key neural circuits that characterize PFF are depicted in Figure \ref{fig:pff_process} and detailed pseudocode (Algorithm 1) for the PFF process can be found in the Appendix.

\noindent
\textbf{Relationship to Contrastive Hebbian Learning:} When designing a network (as we do above), one might notice that the inference process is quite similar to that of a neural system learned under contrastive Hebbian learning (CHL) \cite{movellan1991contrastive}, although there are several significant differences. 
Layer-wise activities in a CHL-based neural system are updated in accordance with the following set of dynamics:
\begin{align}
    \mathbf{z}^\ell(t) &= \mathbf{z}^\ell(t-1) + \beta ( -\mathbf{z}^\ell(t-1) + \mathbf{m}^\ell ) \\
    \mathbf{m}^\ell &= \phi^\ell \Big(\mathbf{W}^\ell \cdot \mathbf{z}^{\ell-1}(t-1) + (\mathbf{W}^{\ell+1})^T \cdot \mathbf{z}^{\ell+1}(t-1) \Big)  \label{eqn:chl_state_update}
\end{align}
where we notice that dynamics do not involve  normalization and the values for layer $\ell$ are integrated a bit differently than in Equation \ref{eqn:ff_state_update}, i.e., neural values change due to leaky Euler integration, where the top-down and bottom-up transmissions are combined to produce a perturbation to the layer rather than propose a new value of the state itself. 

Like CHL, FF and PFF require two phases (or modes of computation) where the signals propagated through the neural system will be used in contrast with one another. 
Given sample $(\mathbf{x},\mathbf{y})$, CHL  entails running the system first in an un-clamped phase (negative phase), where only the input image $\mathbf{x}$ is clamped to the sensory input/bottom layer, followed by a clamped phase, where both $\mathbf{x}$ and its target $\mathbf{y}$ are clamped, i.e., $\mathbf{y}$ is clamped to the output layer (positive phase). At the end of each phase (or inference cycle), the layer-wise activities are recorded and used in a subtractive Hebbian rule to calculate the updates for each synaptic matrix. Note that the positive phase of CHL depends on first running the negative phase. FF and PFF, in contrast, amount to running the positive and negative phases in parallel (with each phase driven by different data), resulting in an overall faster processing time (instead of one inference cycle being conditioned on the statistics of another, the same cycles are now run on both positive or negative data, with opposite objectives \cite{hinton2022forward}, at the same time).

\noindent
\textbf{Relationship to Predictive Coding:} The PFF algorithm integrates the idea of local hypothesis generation from predictive coding (PC) into the inference process by leveraging the representations acquired within the recurrent representation circuit's iterative processing window. Specifically, each layer of the representation circuit, at each time step, becomes the prediction target for each layer of the generative circuit. In contrast, PC  models must leverage a set of feedback synapses to progressively modify their layerwise activities before finally adjusting synaptic values. Furthermore, PFF dynamically modifies  synapses within each processing time step, whereas; typically, PC circuits implement a form of expectation-maximization that, as a result, requires longer stimulus processing windows to learn effective generative models \cite{ororbia2022neural} 
(in this work, the PFF generative circuit learns a good-quality generative model in only $8$-$10$ steps whereas the models of \cite{ororbia2022neural} required at least $50$ steps). 

\begin{figure*}[!t] 
 \centering
     \includegraphics[width=0.75\textwidth]{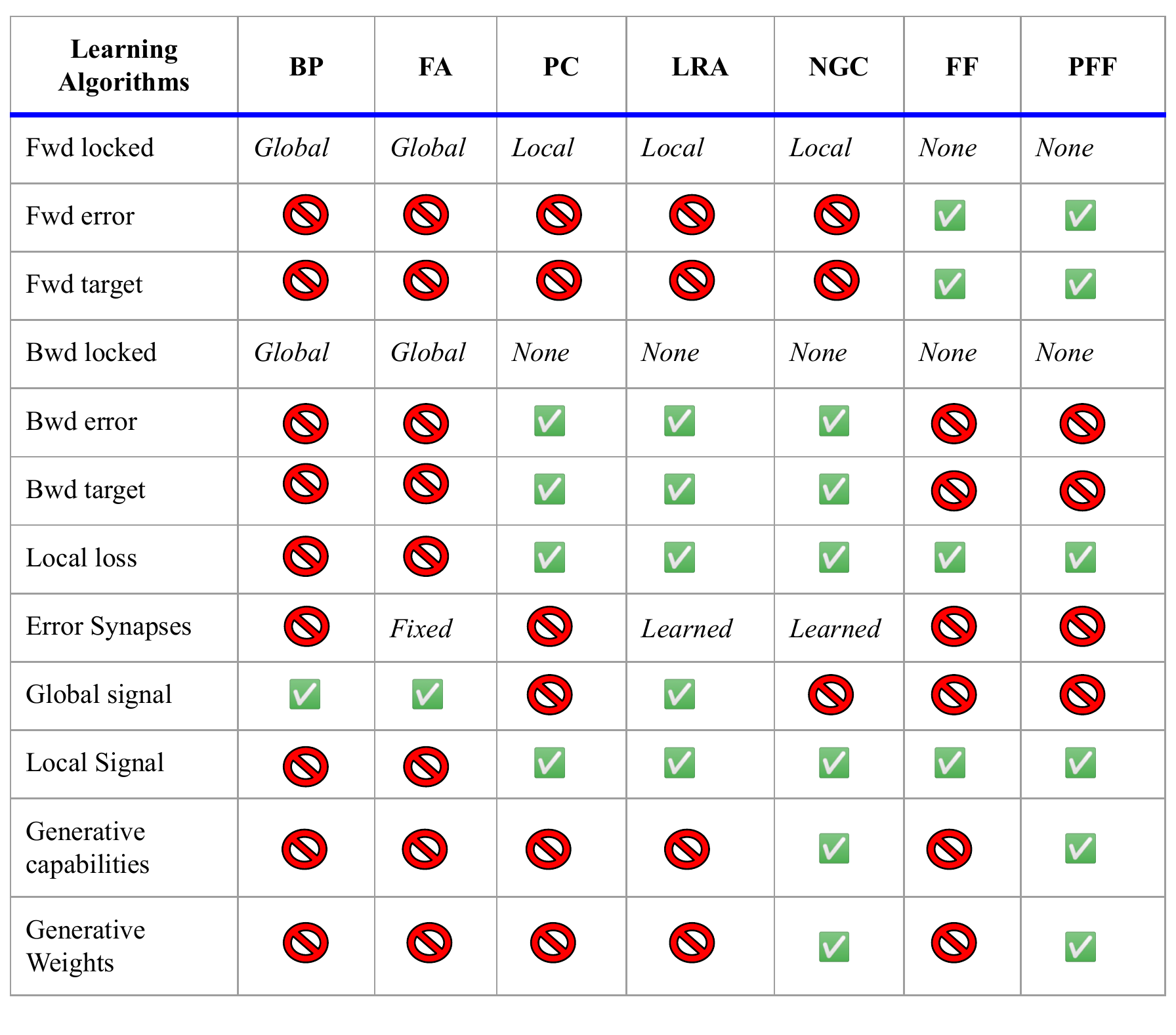}
     \vspace{-0.38cm}
    \caption{Properties of different learning algorithms, i.e., backprop (BP), feedback alignment (FA), predictive coding (PC), local representation alignment (LRA), neural generative coding (NGC), the forward-forward algorithm (FF), and the predictive forward-forward process (PFF).}
    \vspace{-0.6cm}
    \label{fig:algo_properties}
\end{figure*}

\noindent
\textbf{Relationship to Local Learning:}   
It has long been argued that the synapses in the brain are likely to be adjusted according to a local scheme, i.e., only information closest spatially and in time to a target synapse is involved in computing its change in efficacy. Methods that adhere to this biological constraint are referred to as local learning procedures~\cite{ororbia2018biologically,lee2015difference,nokland2016direct,kohan2018error,pmlr-v97-nokland19a,belilovsky2020decoupled,dellaferrera2022error,kohan2023signal,kaiser2020synaptic}, offering a potential replacement for backprop for training deep networks, relaxing one or more of its core constraints (see Figure \ref{fig:algo_properties} for a comparative examination of the key ones across algorithms). 
Desirably, it has even been shown that, empirically, updates from a local scheme can result in improved  generalization~\cite{lee2015difference,ororbia2018biologically}, even with temporal data \cite{mali2021investigating} and discrete signals \cite{mali2021empirical}. There have been many efforts in designing biologically-plausible local learning algorithms, such as contrastive Hebbian learning (mentioned above) \cite{movellan1991contrastive}, contrastive divergence for learning harmoniums (restricted Boltzmann machines) \cite{hinton2002training}, the wake-sleep algorithm for learning Helmholtz machines \cite{hinton1995wake}, and algorithms such as equilibrium propagation \cite{scellier2017equilibrium}.  
Other efforts directly integrate local learning into the deep learning pipeline, e.g., kickback~\cite{balduzzi2015kickback} and decoupled neural interfaces~\cite{jaderberg2016decoupled}. 
It is worth pointing out that PFF bears similarity to the wake-sleep algorithm, which entails learning a generative model jointly with an inference (recognition) model. However, the wake-sleep algorithm suffers from instability, given that the recognition network could be damaged by random fantasies produced by the generative network and the generative network could itself be hampered by the low-quality representation capability of the inference network (motivating variations such as reweighted wake-sleep \cite{bornschein2014reweighted}). PFF instead aims to learn the generative model given the representation circuit, using locally-adapted neural activities as a guide for the synthesization process rather than randomly sampling the generative model to create teaching signals for the recognition network (which would potentially distract its optimization with nonsensical signals).

\section{Experiments}
\label{sec:experiments}

This section describes the simulations/experiments designed to test the capability of the proposed PFF procedure. We leverage several image datasets to quantitatively evaluate PFF's classification ability (in terms of test-set error) and qualitatively evaluate its generative capability (in terms of visual inspection of sample reconstruction and pattern synthesization). The PFF process (PFF-RNN) is compared to several baselines, including the $K$-nearest neighbors algorithm (with $K=4$, or 4-KNN), the original recurrent network trained with the original FF algorithm \cite{hinton2022forward} (FF-RNN), and two backprop-based models, i.e., a feedforward network that uses backprop to adjust all of its internal synapses (BP-FNN) and the same network but one that only adjusts the top-most softmax/output layer parameters and fixes the hidden layer synaptic parameters (Rnd-FNN). Both backprop-based networks are trained to minimize the categorical cross-entropy of each dataset's provided labels. The partially-trained model, i.e., the Rnd-FNN, serves as a sort of lower bound on the generalization ability of a neural system, given that it is possible to obtain respectable classification performance with only random hidden feature detectors (note that a neural credit assignment algorithm should not perform worse than this).

\noindent
\textbf{Datasets:} We experiment with several (gray-scale) image collections, i.e., the MNIST, Kuzushiji-MNIST (K-MNIST), Fashion MNIST (F-MNIST), Not-MNIST (N-MNIST), and Ethiopic (Et-MNIST) databases. All databases contain $28\times28$ images from $10$ different categories (see Appendix for details). 

\noindent
\textbf{Simulation Setup:} All simulated models were constrained to use similar architectures to ensure a fair comparison. All networks for all neural-based learning algorithms contained two hidden layers of $2000$ neurons (much akin to the FF models in \cite{hinton2022forward}), with initial synaptic weight values selected according to the random orthogonal initialization scheme \cite{saxe2013exact} (using singular value decomposition). Once any given algorithm calculated adjustment values for the synapses, parameters were adjusted using the Adam adaptive learning rate \cite{kingma2014adam} with mini-batches containing $500$ samples. 
Both FF and PFF were set to use a threshold value of $\theta_z = 10.0$, and PFF was set to use $20$ latent variables (i.e., $\mathbf{z}_s \in \mathcal{R}^{20 \times 1}$), representation noise $\sigma^\ell = 0.05$, and generative noise $\sigma_z = 0.025$. 

\begin{table*}[!t]
\caption{Classification generalization results for systems trained under different learning algorithms (except for 4-KNN, which is a non-parametric learning baseline model). Measurements of mean and standard deviation are for five experimental trial runs.}
\centering
\begin{tabular}{l|c|c|c|c|c}
 & \textbf{MNIST} & \textbf{K-MNIST} & \textbf{F-MNIST} & \textbf{N-MNIST} & \textbf{Et-MNIST}\\
\textbf{Model} & \textbf{Test Error} (\%) & \textbf{Test Error} (\%) & \textbf{Test Error} (\%) & \textbf{Test Error} (\%) & \textbf{Test Error} (\%)\\
\hline
4-KNN & $2.860 \pm 0.000$ & $7.900 \pm 0.000$ & $14.030 \pm 0.00$ & $7.720 \pm 0.000$ & $2.120 \pm 0.000$\\
Rnd-FNN & $3.070 \pm 0.018$ & $14.070 \pm 0.189 $ & $17.100 \pm 0.102$ & $9.916 \pm 0.135$ & $1.650 \pm 0.040$\\
BP-FNN & $\mathbf{1.300 \pm 0.023}$ & $6.340 \pm 0.202$ & $10.720 \pm 0.014$ & $5.574 \pm 0.09$ & $\mathbf{0.392 \pm 0.02}$\\
FF-RNN  & $1.320 \pm 0.100$ & $6.590 \pm 0.420$ & $10.750 \pm 0.060$ & $5.380 \pm 0.040$ & $0.470 \pm 0.050$\\ 
PFF-RNN & $1.340 \pm 0.010$ & $\mathbf{6.250 \pm 0.060}$ & $\mathbf{10.400 \pm 0.010}$ & $\mathbf{5.270 \pm 0.101}$ & $0.451 \pm 0.070$\\ 
\hline
\end{tabular}
\vspace{-0.5cm}
\label{results:classify}
\end{table*} 

\begin{figure*}[!t]
     \centering
     \begin{subfigure}[t]{0.195\textwidth}
         \centering
         \includegraphics[width=\textwidth]{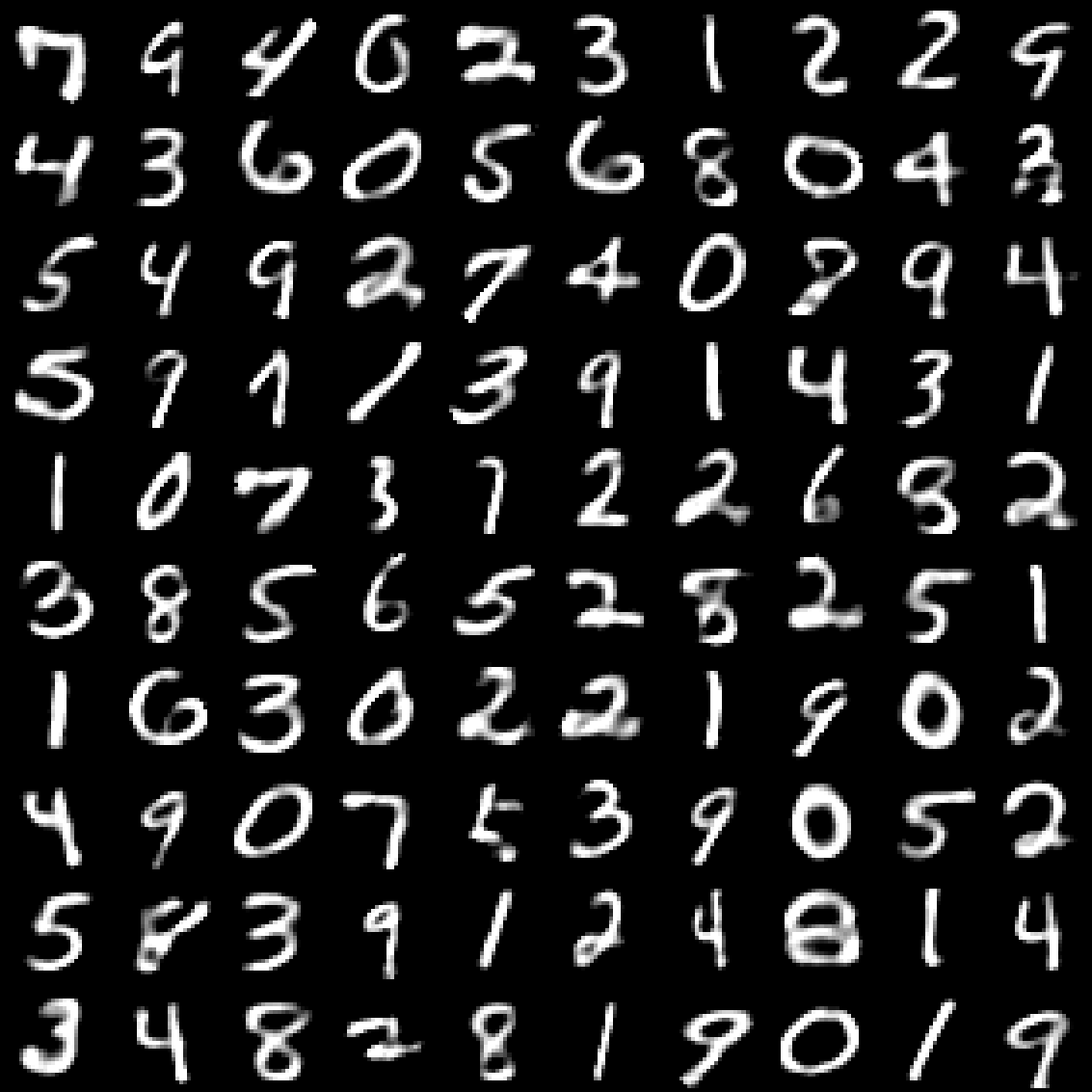}
         \caption{MNIST recon.}
         \label{fig:mnist_recon}
     \end{subfigure}
     \begin{subfigure}[t]{0.195\textwidth}
         \centering
         \includegraphics[width=\textwidth]{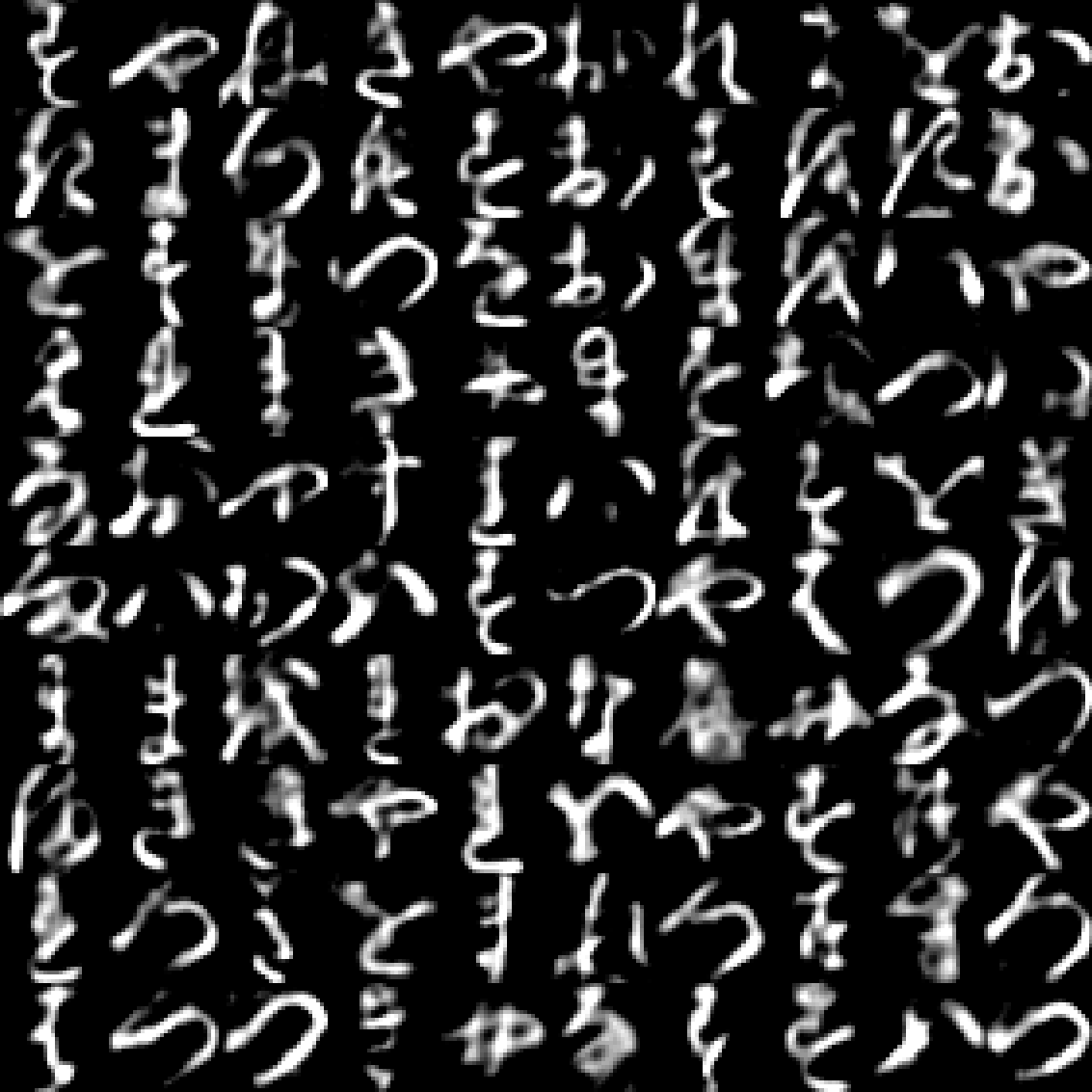}
         \caption{K-MNIST recon.}
         \label{fig:kmnist_recon}
     \end{subfigure}
    \begin{subfigure}[t]{0.195\textwidth}
         \centering
         \includegraphics[width=\textwidth]{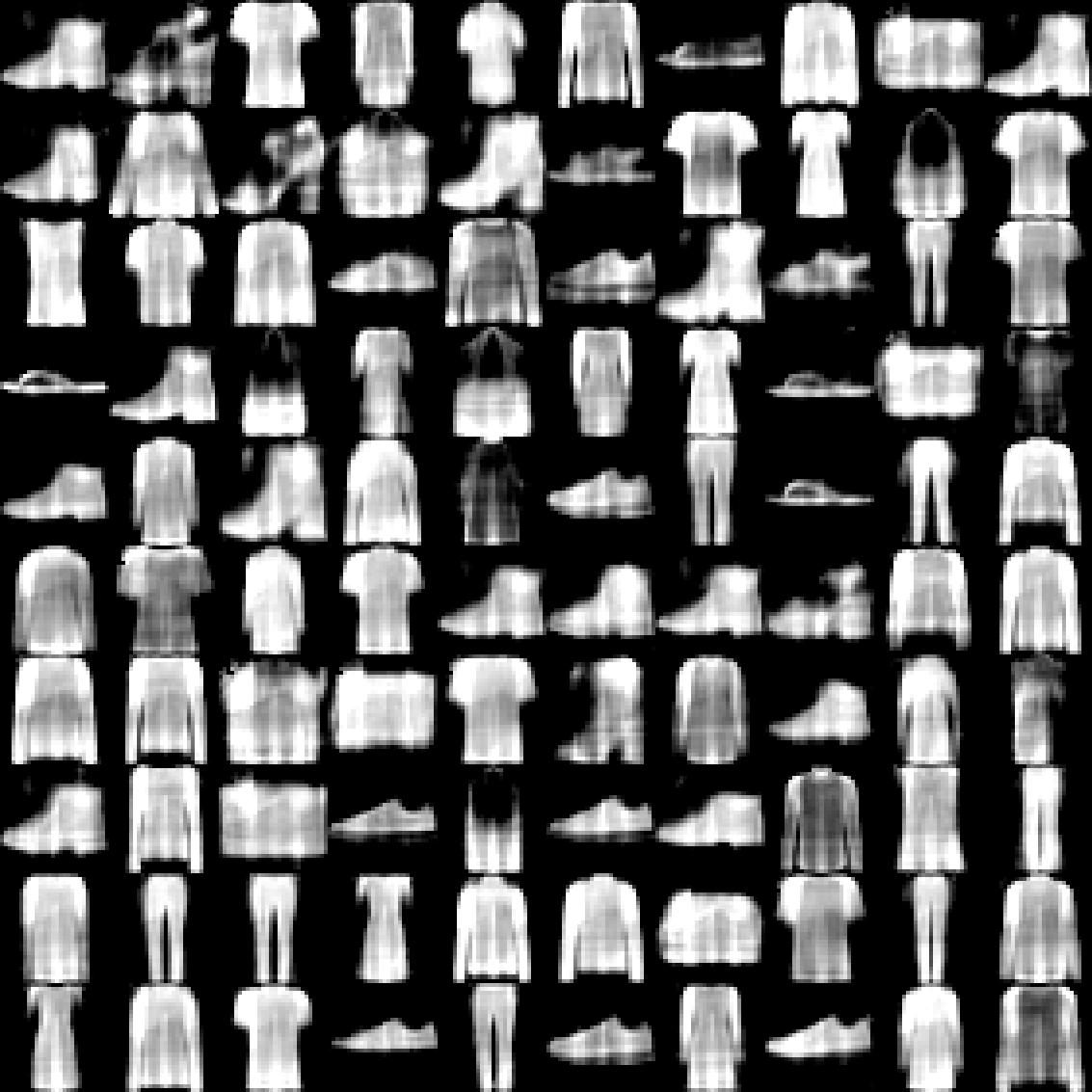}
         \caption{F-MNIST recon.}
         \label{fig:fmnist_recon}
     \end{subfigure}
     \begin{subfigure}[t]{0.195\textwidth}
         \centering
         \includegraphics[width=\textwidth]{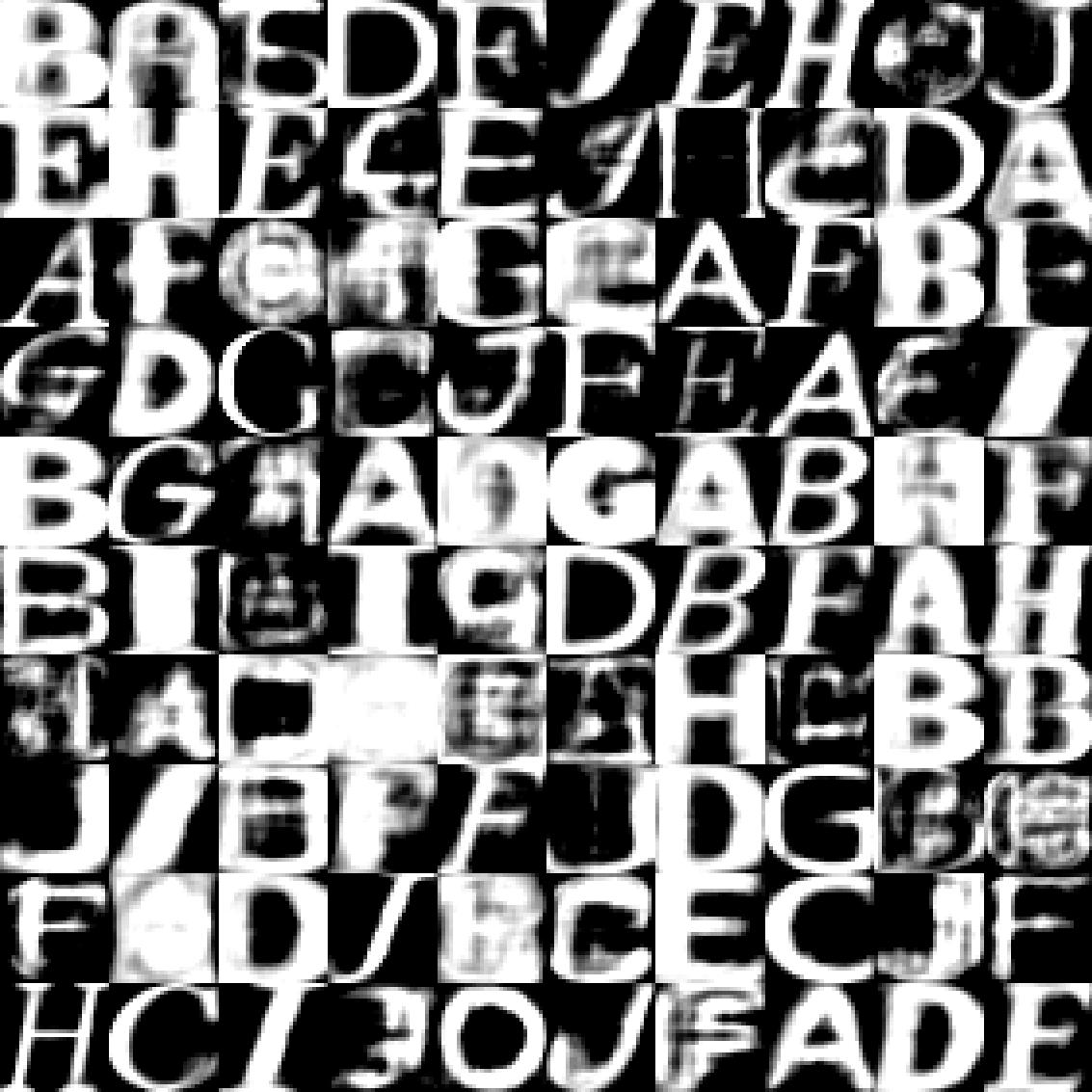}
         \caption{N-MNIST recon.}
         \label{fig:nmnist_recon}
     \end{subfigure}
     \begin{subfigure}[t]{0.195\textwidth}
         \centering
         \includegraphics[width=\textwidth]{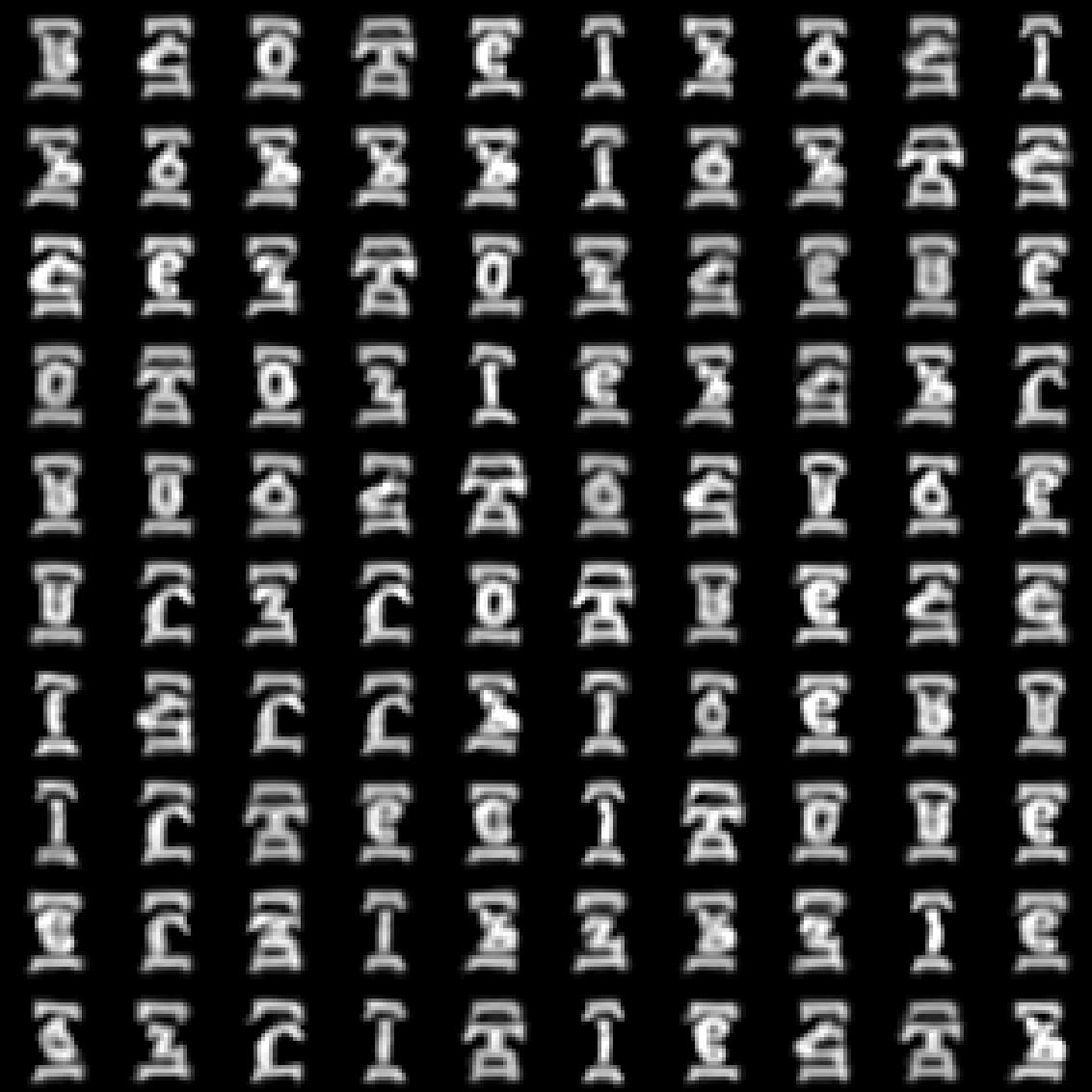}
         \caption{E-MNIST recon.}
         \label{fig:emnist_recon}
     \end{subfigure} \\
    \begin{subfigure}[t]{0.195\textwidth}
         \centering
         \includegraphics[width=\textwidth]{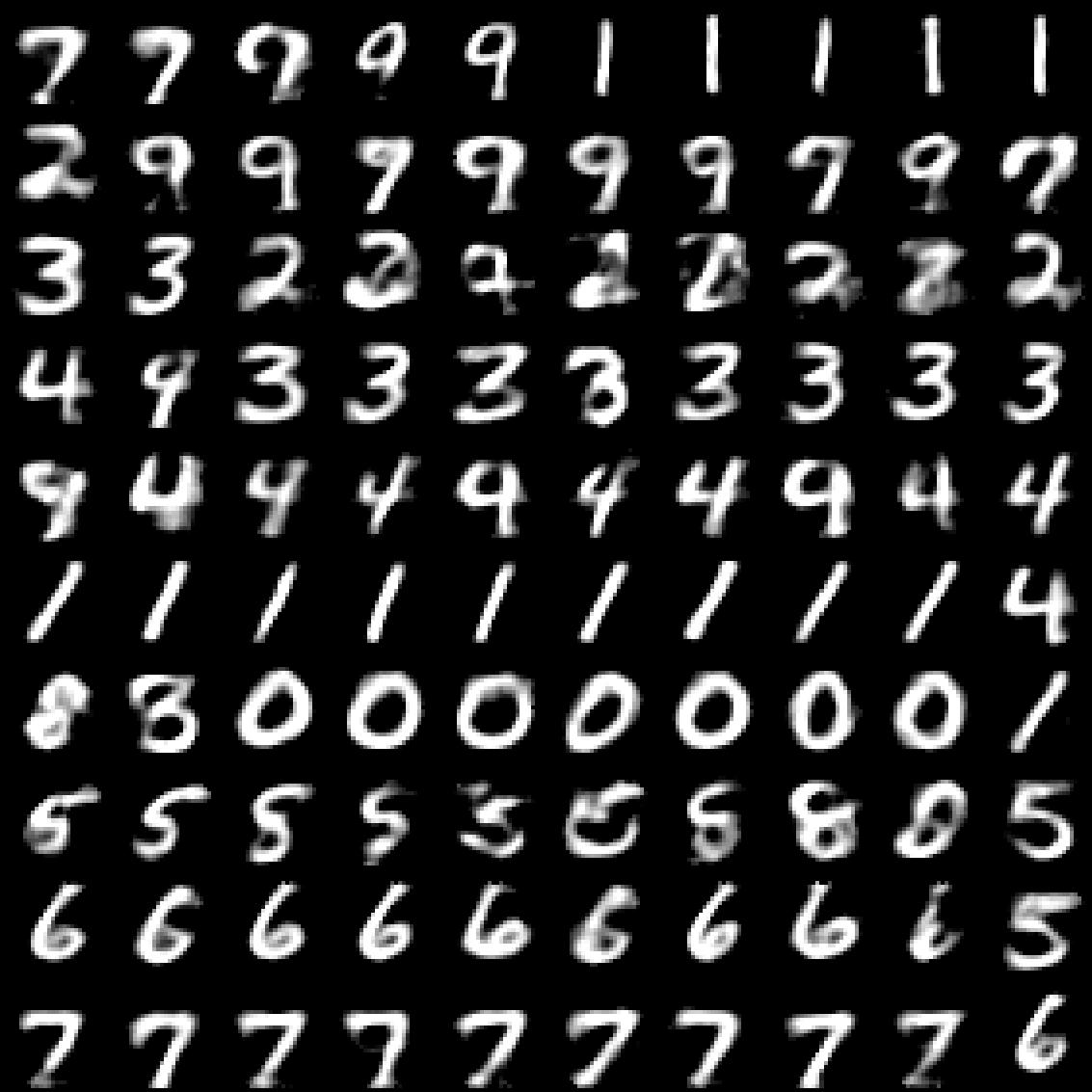}
         \caption{MNIST synthesis.}
         \label{fig:mnist_samples}
     \end{subfigure}
     \begin{subfigure}[t]{0.195\textwidth}
         \centering
         \includegraphics[width=\textwidth]{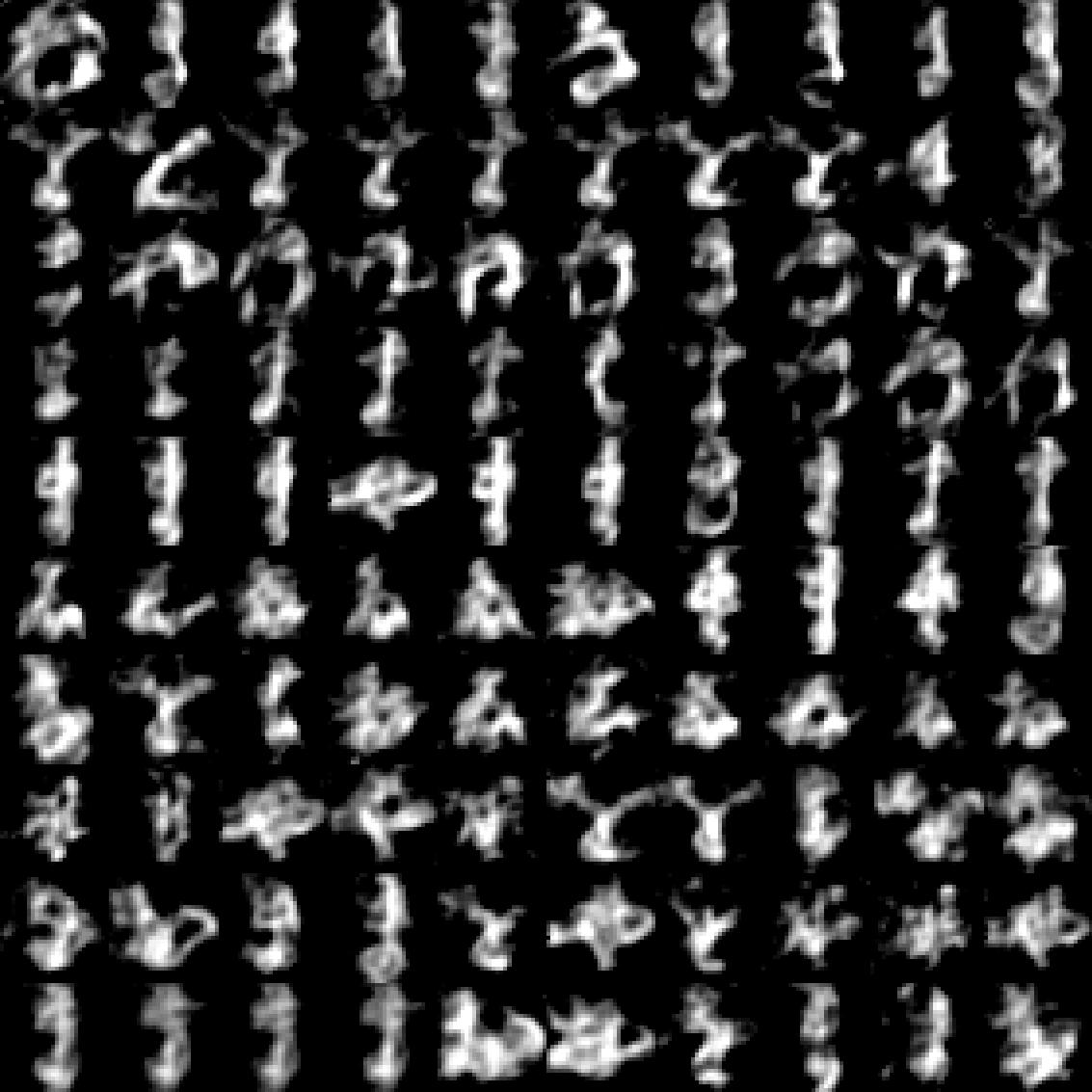}
         \caption{K-MNIST synthesis.}
         \label{fig:kmnist_samples}
     \end{subfigure}
    \begin{subfigure}[t]{0.195\textwidth}
         \centering
         \includegraphics[width=\textwidth]{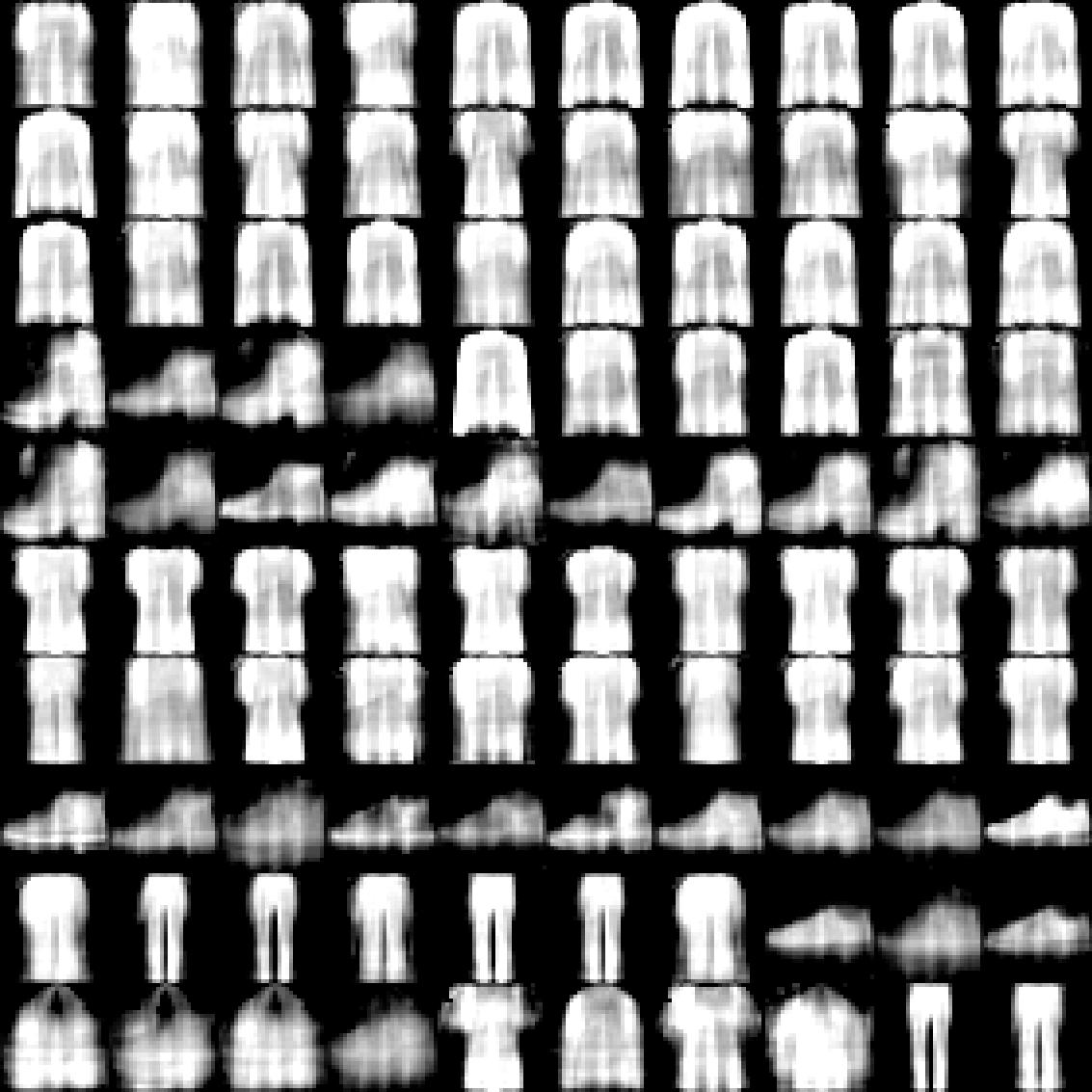}
         \caption{F-MNIST synthesis.}
         \label{fig:fmnist_samples}
     \end{subfigure}
     \begin{subfigure}[t]{0.195\textwidth}
         \centering
         \includegraphics[width=\textwidth]{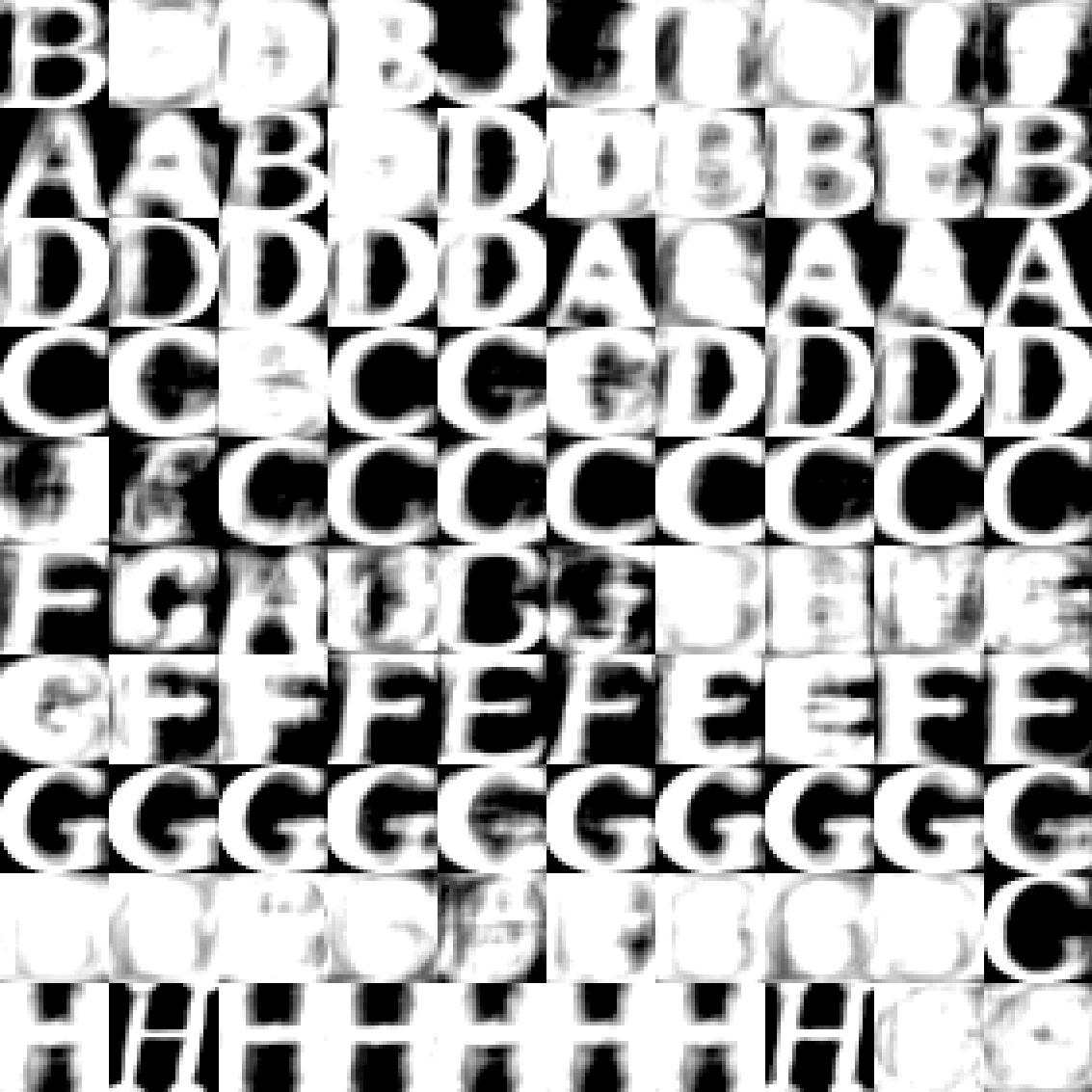}
         \caption{N-MNIST synthesis.}
         \label{fig:nmnist_samples}
     \end{subfigure}
     \begin{subfigure}[t]{0.195\textwidth}
         \centering
         \includegraphics[width=\textwidth]{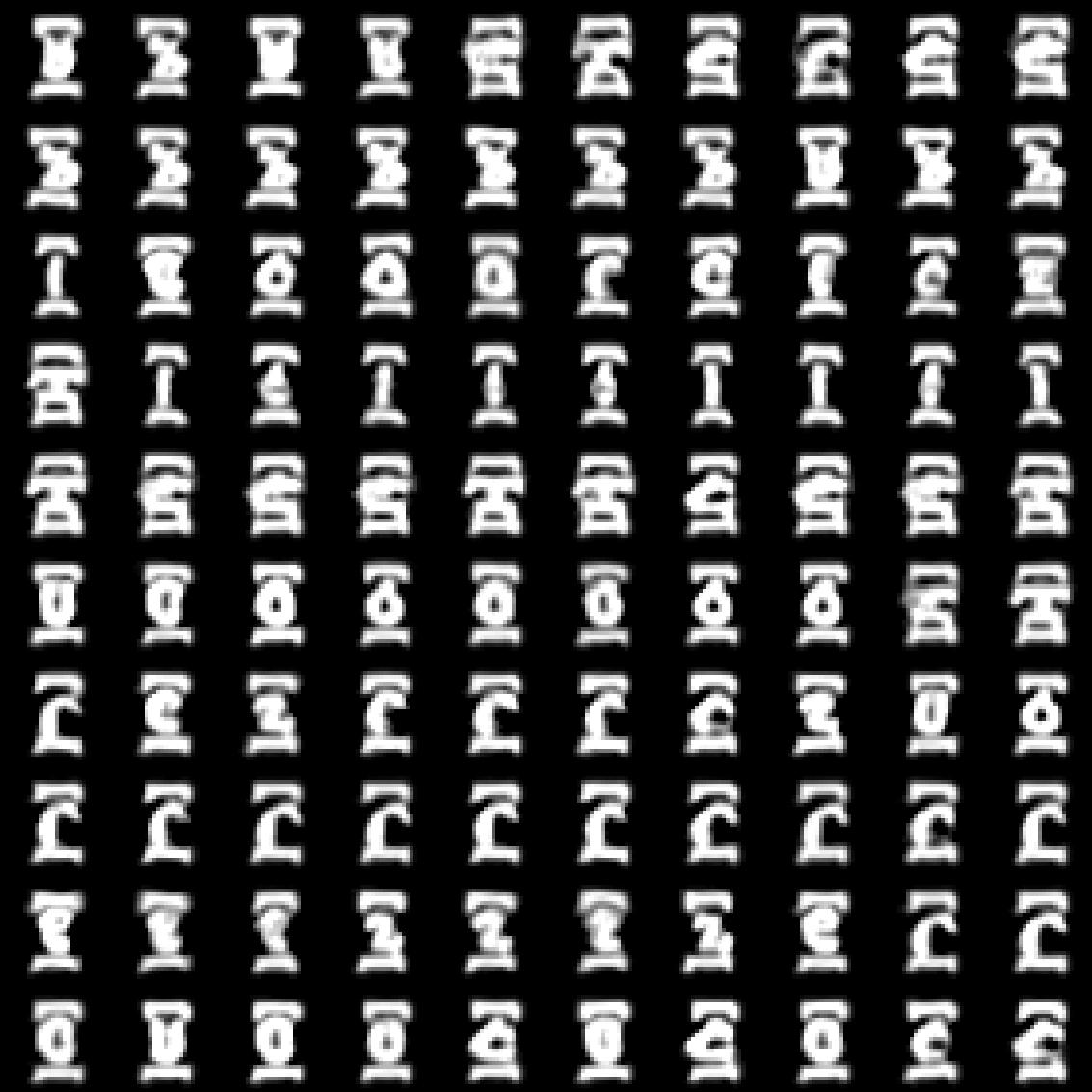}
         \caption{E-MNIST synthesis.}
         \label{fig:emnist_samples}
     \end{subfigure} \\
     \begin{subfigure}[t]{0.195\textwidth}
         \centering
         \includegraphics[width=\textwidth]{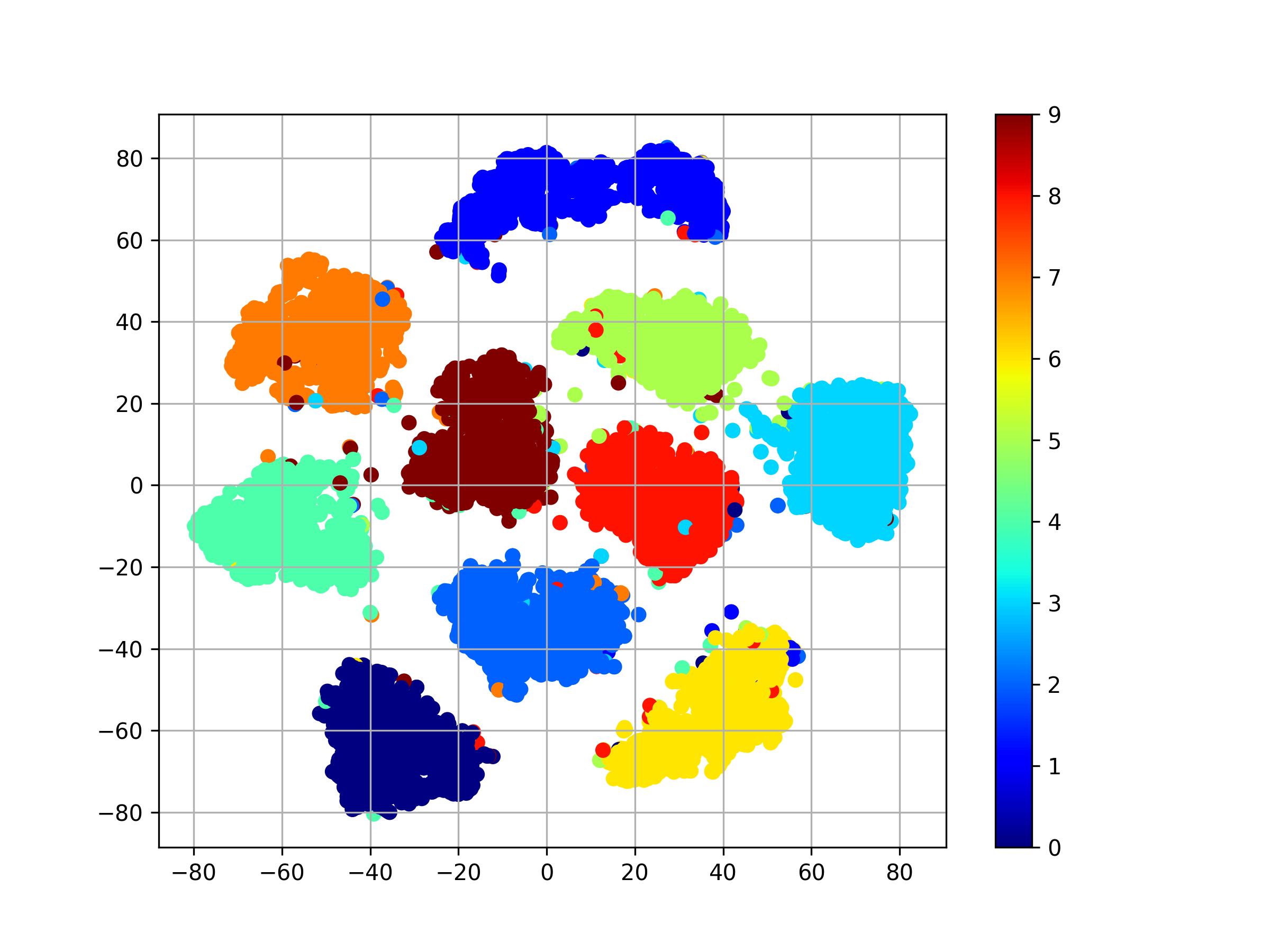}
         \vspace{-0.75cm}
         \caption{MNIST latents.}
         \label{fig:mnist_lat}
     \end{subfigure}
     \begin{subfigure}[t]{0.195\textwidth}
         \centering
         \includegraphics[width=\textwidth]{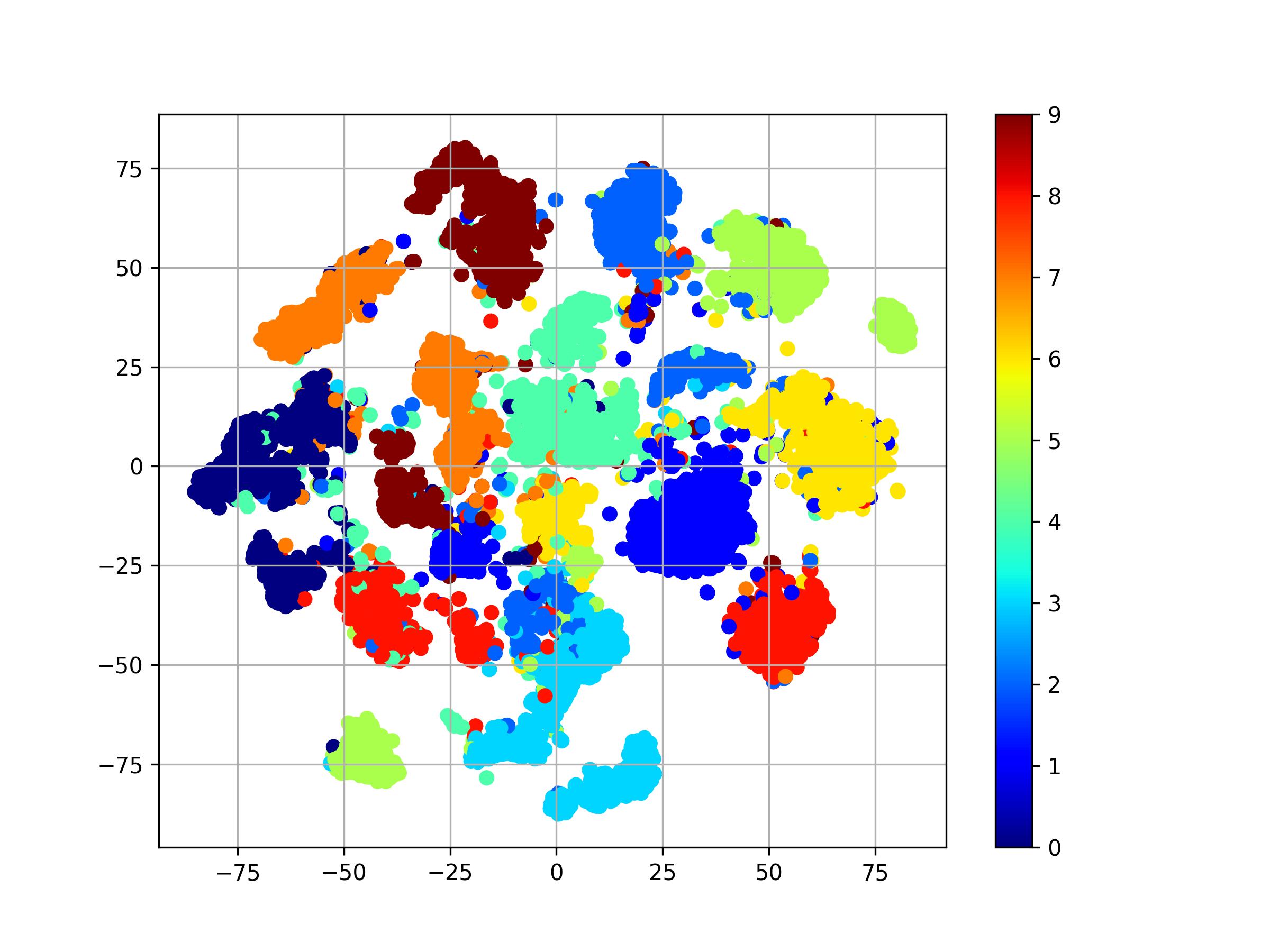}
         \vspace{-0.75cm}
         \caption{K-MNIST latents.}
         \label{fig:kmnist_lat}
    \end{subfigure}
    \begin{subfigure}[t]{0.195\textwidth}
         \centering
         \includegraphics[width=\textwidth]{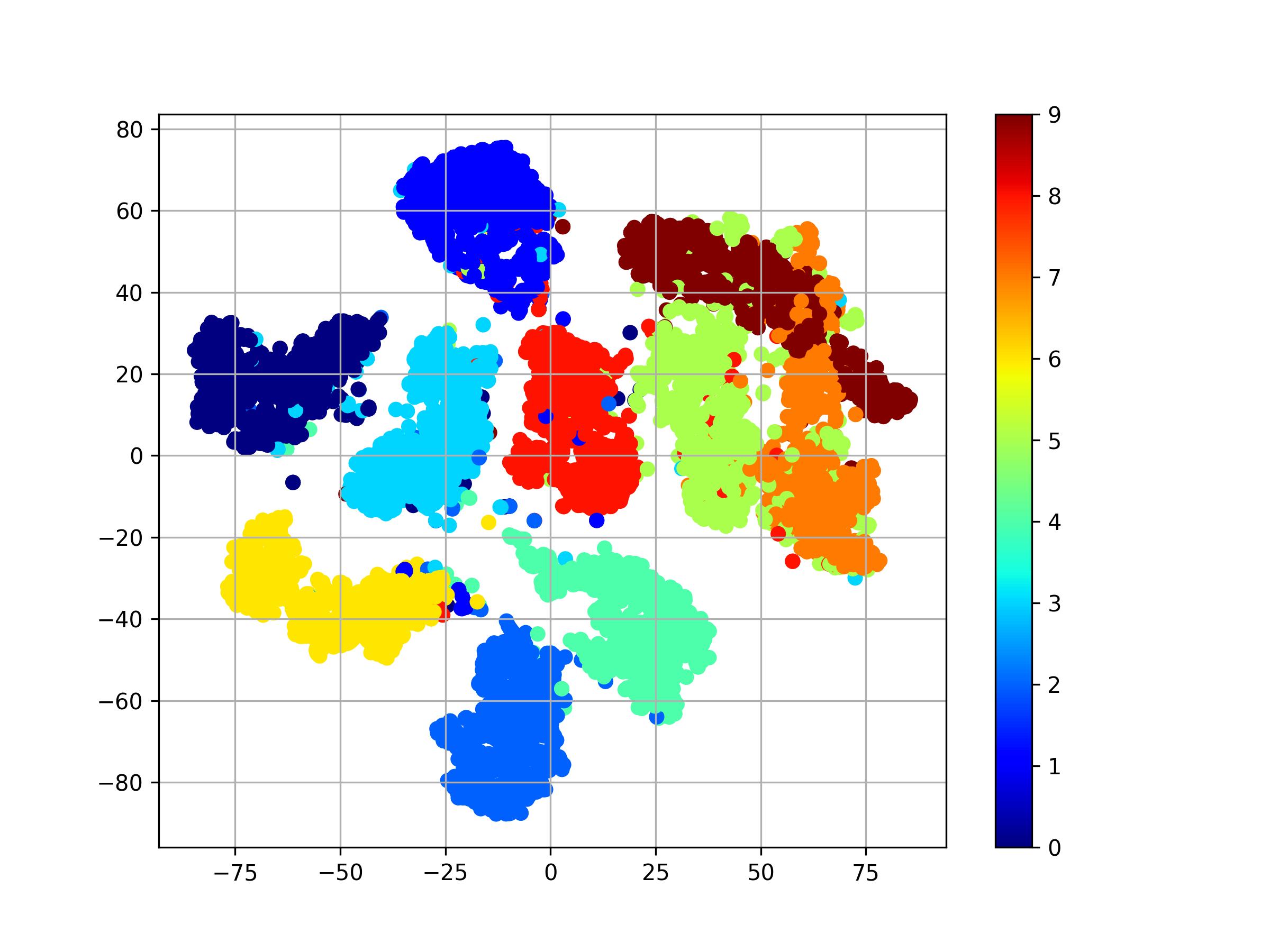}
         \vspace{-0.75cm}
         \caption{F-MNIST latents.}
         \label{fig:fmnist_lat}
     \end{subfigure}
     \begin{subfigure}[t]{0.195\textwidth}
         \centering
         \includegraphics[width=\textwidth]{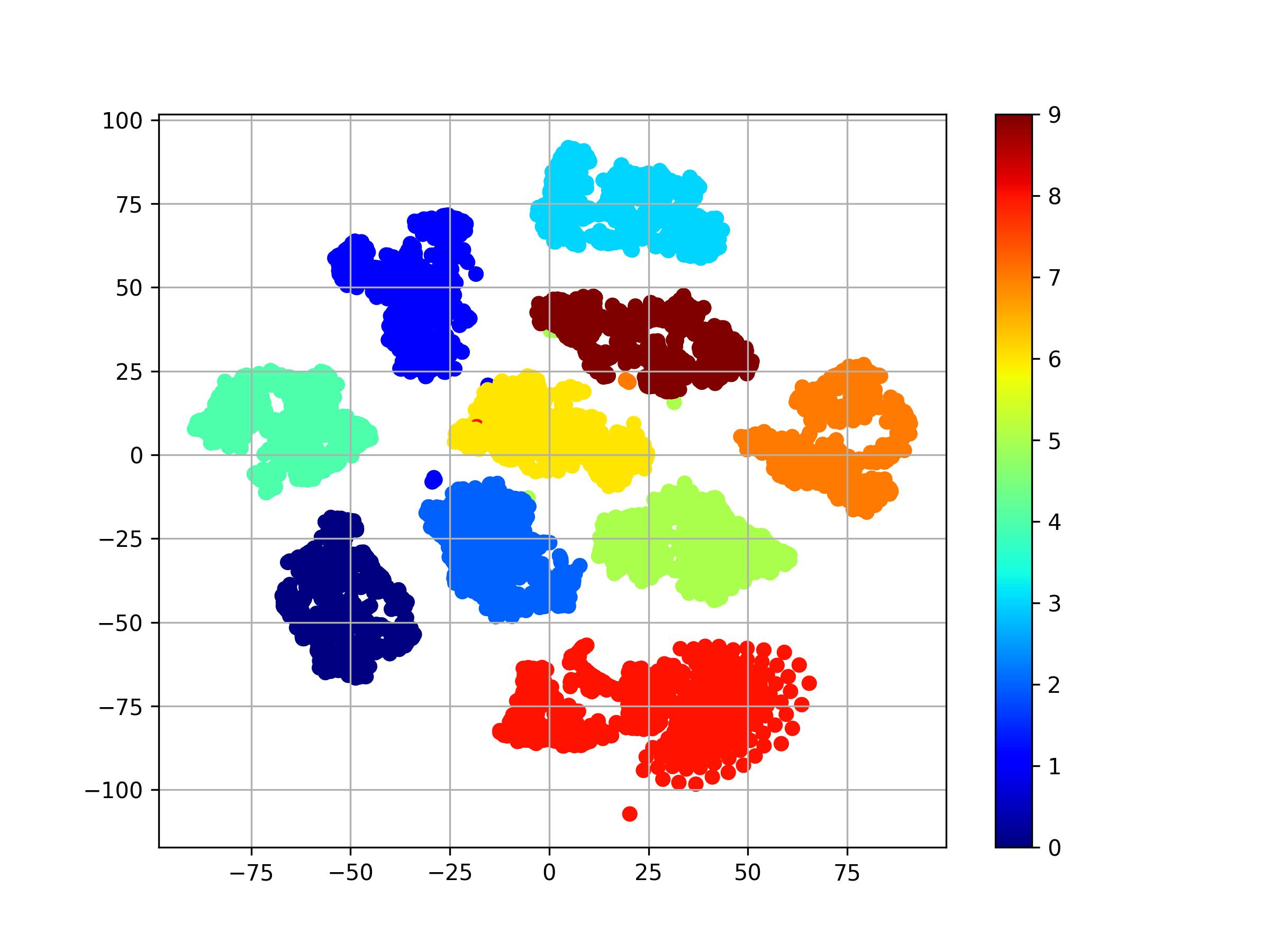}
         \vspace{-0.75cm}
         \caption{N-MNIST latents.}
         \label{fig:nmnist_lat}
     \end{subfigure}
     \begin{subfigure}[t]{0.195\textwidth}
         \centering
         \includegraphics[width=\textwidth]{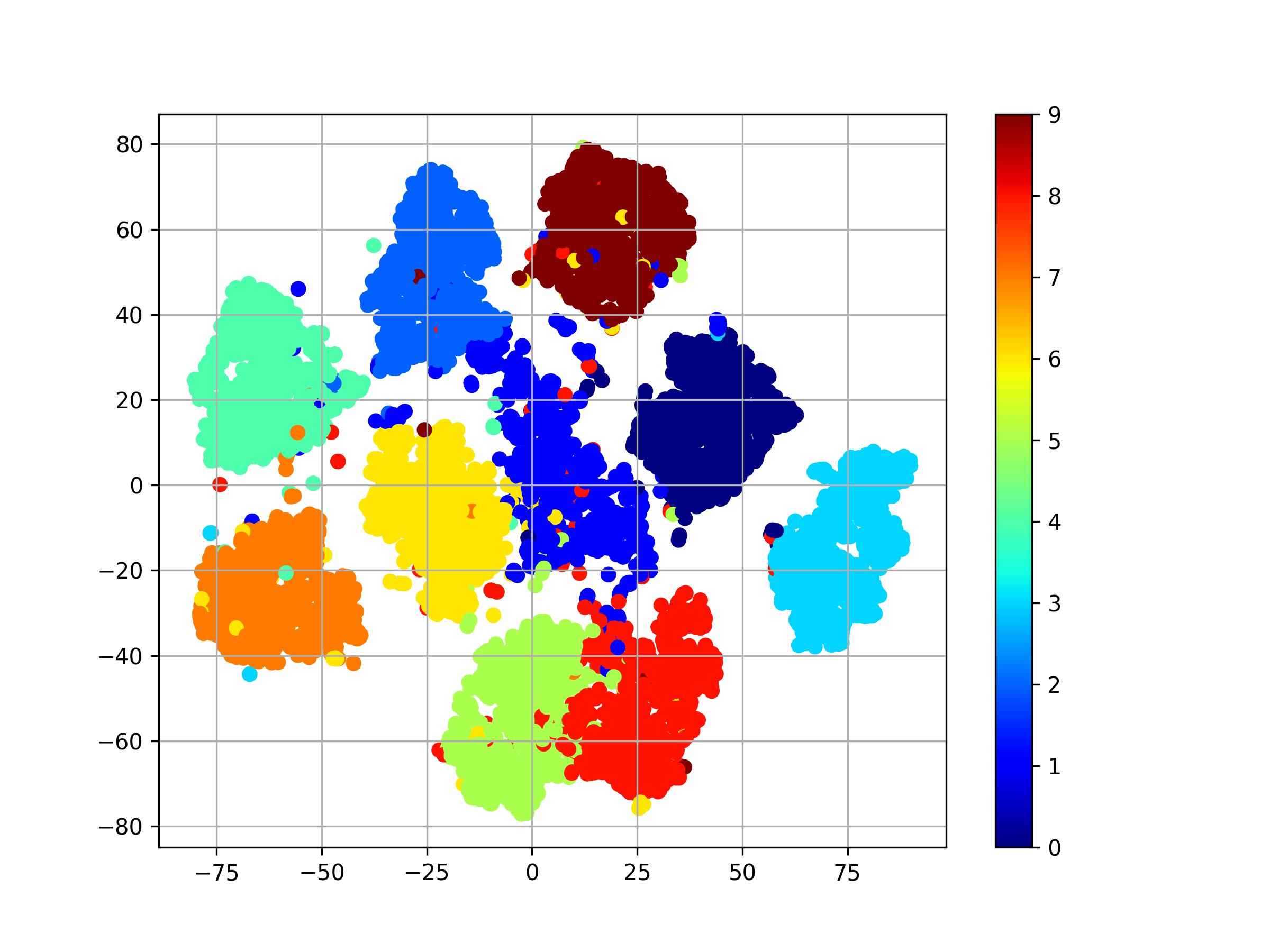}
         \vspace{-0.75cm}
         \caption{E-MNIST latents.}
         \label{fig:emnist_lat}
     \end{subfigure}
        \caption{Model reconstructed patterns (``recon'', top row) and generated samples (``synthesis'', middle row) for each natural image database. In the bottom row, visualizations of the top-most latent activity layer of the PFF-RNN's generative circuit for each dataset.}
        \label{fig:model_samples}
        \vspace{-0.5cm}
\end{figure*} 

\noindent
\textbf{Discussion:} 
Observe in Table \ref{results:classify} that the PFF procedure performs well for the models simulated in this study, reaching a top/good-quality classification error of about $1.34$\% on MNIST, nearly reaching that of the well-tuned classifier BP-FNN ($1.30$\%). Notably, the PFF-RNN model outperforms BP-FNN on $3$ of the (arguably harder) datasets. Both FF and PFF outperform the lower-bound baselines, i.e., 4-KNN and Rnd-FNN, indicating that they acquire feature detectors that facilitate good discriminative performance. 

Qualitatively, in Figure \ref{fig:model_samples} (top row), observe that PFF learns a good-quality reconstruction model and generative model of the image inputs. The reconstructed symbols (e.g., digits, Kanji characters) are excellent, and the image samples for all cases exhibit some variety across the categories (albeit a bit blurry). Note that to sample from the PFF's directed generative model, as mentioned earlier in the section ``The Generative Circuit'', 
we retro-fit a GMM to samples of the latent variable $\mathbf{z}_s$, optimizing a GMM (with $10$ components) via expectation-maximization. Finally, in the bottom row of Figure \ref{fig:model_samples}, we visualize the latent space induced by the top-most activity layer of  PFF's generative circuit, observing that in all cases, class-centric clusters desirably emerge. In the Appendix, we visually inspect the receptive fields (of synapses closest to the sensory input layer) of both PFF circuits and discuss in detail the computational benefits afforded by the PFF process. 

\noindent
\textbf{Computational Benefits of the Predictive Forward-Forward Algorithm: } 
From a hardware efficiency point-of-view, the PFF algorithm, much like the FF procedure\footnote{We remark that this would also apply to other forward-only learning algorithms \cite{kohan2018error,kohan2023signal}.}, is a potentially promising candidate for implementation in analog and neuromorphic hardware. It is the fact that FF and PFF only require forward passes\footnote{FF requires, per simulated time step, two forward passes and PFF requires three forward passes.} to conduct inference and synaptic adjustment that creates this possibility, given that these algorithms require no distinct separate computational pathway(s) needed for transmitting teaching signals (required by backprop \cite{rumelhart1988backprop} and feedback alignment methods \cite{lillicrap2014random,nokland2016direct}) or even error messages (required by predictive coding \cite{rao1999predictive,bastos2012canonical,ororbia2022neural}, representation alignment \cite{ororbia2018biologically}, and target propagation processes \cite{lee2015difference}). Desirably, this means that  no specialized hardware is needed for calculating derivatives (which is needed for activation functions -- this means that even implementations of PFF with discrete and non-differentiable stochastic functions, e.g., sampling the Bernoulli distribution, is possible/viable) nor for maintaining and adjusting in memory separate feedback transmission synapses (that often require different adjustment rules \cite{ororbia2018biologically,ororbia2018continual}). 

Note effort is still be needed to properly reformulate PFF and FF to be completely compatible with neuromorphic/analog chips, given that both procedures still utilize matrix multiplications to simulate information transmission across neural regions. Spiking neural systems (which are suited for simulation on such hardware), in contrast, require activities to be implemented as voltages (coupled to spike-response models) and synapses as conductances which allows neural charges (per unit time) to be added. However, an approach similar to that taken by spiking neural coding \cite{ororbia2019spiking} could be adopted to transform FF/PFF natural to leveraging the sparse computations inherent to spike trains. However, given that both FF and PFF are forward-pass only credit assignment algorithms, analog-to-digital conversion processes would no longer be required to facilitate a backward pass (as required by backprop) needed to calculate numerical gradients. In addition, the lateral synaptic weights that further induce structural sparsity in PFF's representation circuit would be particularly useful to implement for hardware that directly supports sparse computations. 
Finally, we remark that PFF, as noted in the main paper, is faster than forward-only algorithms based on contrastive Hebbian learning setups \cite{movellan1991contrastive,hinton2002training,scellier2017equilibrium}, given that it does not require its negative phase computation to be conditioned on the statistics of the positive one (or vice versa). 

\noindent 
\textbf{Intertwined Inference-and-Learning and Mortal Computation: } 
Another important point to make is that the PFF algorithm, and future algorithms like it, provide a computational framing referred to in \cite{hinton2022forward} as ``mortal computation'', or systems where the software and hardware are no longer distinctly and clearly separated. In some sense, the notion of mortal computation is in the same spirit as \emph{\textbf{intertwined inference-and-learning}} \cite{ororbia2017learning,ororbia2018biologically,ororbia2018continual,ororbia2022neural}, which refers to the fact that neurobiological learning and inference in the brain are not really two completely distinct and separate processes but rather complementary ones which depend on one another, even relying and depending on the very structure of the neural circuits that conduct them. Intertwined inference-and-learning, with predictive coding/processing and contrastive Hebbian learning as its key exemplars, means that without the inference process, there exists no synaptic adjustment, and without synaptic adjustment, inference becomes meaningless (as knowledge would not be encoded into any form of longer-term memory). 
Mortal computation could be considered to be the next logical extension and consequence of intertwined inference-and-learning, given that the knowledge encoded in simulated synaptic weights/values (as well as the computation that they ultimately conduct/support) ``dies'' or disappears when the hardware fails/dies.

Embracing the notion of mortal computation, which comes with the trade-off that a computational program/simulation of neural computation can no longer be naturally copied to millions of computers/devices (to take advance of high-performance computing systems), could bring about substantial savings in energy expenditure, which has increasingly become a concern in artificial intelligence (AI) research \cite{amodei_hernandez_2018,brevini2020black,strubell2019energy,schwartz2020green}, i.e., how may we avoid designing intelligent systems that rapidly escalate computational and carbon costs, or \emph{Red AI} \cite{schwartz2020green}, in favor of \emph{Green AI} approaches. 
Ultimately, this means that crafting intertwined learning-and-inference approaches that account for their manifestation in a target physical structure, e.g., form and design of the hardware, could prove invaluable going forward in statistical learning research, providing a means to run massive neural systems (that contain trillions of synapses) while only consuming a few watts of energy. Notably, this might lead to the design of neural systems that are aware and adapt based on the state of the very hardware they run on, emulating homeostatic constraints that define biological learning and inference. 
However, much work remains in designing algorithms that run efficiently on hardware of which the precise details are largely unknown, and investigating how processes such as PFF and FF scale to larger neural systems could be an important step for doing so.

\noindent 
\textbf{On Self-Generated Negative Samples: } One of the most important elements of PFF is its integrated, jointly-learned predictive/generative neural circuit. This generative model, as we have shown in the main paper, is capable not only of high-quality reconstruction of the original input patterns but also of synthesizing data by sampling its latent prior $P(\mathbf{z}_s)$, which we designed to be a Gaussian mixture model due to the highly multi-modal latent space induced by its top-most neural activities. An important future direction to explore with FF-based biological credit assignment algorithms is the examination of different schedules of the positive and negative phases in contrast to the simultaneous ones used in \cite{hinton2022forward} and this work. Specifically, an important question to answer is: \emph{To what degree can the positive and negative phases be separated while still facilitating stable and effective local adaptation of a neural circuit's synapses?} 
Ours and other future efforts should explore this question in depth leveraging PFF's jointly generative/predictive circuit -- one starting point could entail extending the generative circuit to be conditioned on a label vector $\mathbf{y}$ and running it at certain points in training (e.g., after a pass through $M$ data samples or batches) to synthesize several batches of patterns and incorrectly mapping them to clearly incorrectly label inputs (i.e., choose $\mathbf{y}$ to get a particular sample of a certain class and then purposely automatically select an incorrect label knowing the originally chosen one). 

Crafting schemes such as the one sketched above could also facilitate the development of interesting self-supervised approaches to continual learning, allowing a PFF system to use a form of internal, self-induced (memory) replay to refresh its knowledge on older data points, potentially offering an efficient means of combating the grand challenge of catastrophic forgetting in artificial neural networks. As mentioned in \cite{hinton2022forward}, if the positive and negative phases could be separated or be made more distant from each other, one could potentially model/investigate the effects of ``severe sleep deprivation'' by eliminating the negative phase updates for a period of time. Finally, from a theoretical point-of-view, with implications for computational neuroscience, cognitive science, and statistical learning, credit assignment through the PFF process offers an important stepping stone towards developing a unified theoretical framework for both forward-forward based learning and predictive processing.

\section{Conclusion}
\label{sec:conclusion}

In this work, we proposed the predictive forward-forward (PFF) process for dynamically adjusting the synaptic efficacies of a recurrent neural system that jointly learns how to classify, reconstruct, and synthesize data samples without backpropagation of errors. Our model and credit assignment procedure integrates elements of the forward-forward algorithm, such as its local synaptic adaptation rule based on goodness and contrastive learning, with a novel, simple form of lateral synaptic competition as well as aspects of predictive coding, such as its local error Hebbian manner of adjusting generative synaptic weights. Our results indicate that the PFF learning process offers a promising brain-inspired, forward-only, backprop-free form of credit assignment for neural systems. PFF could prove useful not only for crafting more biologically faithful neural models but also as a candidate circuit/module for neural-based cognitive architectures such as the COGnitive Neural GENerative system (CogNGen) \cite{Ororbia2022cogsci}, which seek to explain how higher-level cognitive function emerges from lower-level neural cellular activity.

\bibliographystyle{acm}
\bibliography{ref}

\newpage
\section*{Appendix}

\subsection*{The PFF Learning Process}
\label{sec:pff_algorithm}

In Algorithm \ref{alg:inference}, we present the full formal specification of the PFF inference-and-learning process. Note that the notation used in the pseudocode adheres to the same notation defined/presented in the main paper. 

\subsection*{On Lateral Competition Synapses}

In the PFF algorithm, we introduced a learnable form of lateral inhibition and self-excitation into the representation circuit's dynamics as follows (reproduced below for convenience):
\begin{align}
    \mathbf{z}^\ell(t) =  \beta \Big( \phi^\ell \big( &\mathbf{W}^\ell \cdot \text{LN}( \mathbf{z}^{\ell-1}(t-1) ) + \mathbf{V}^\ell \cdot \text{LN}( \mathbf{z}^{\ell+1}(t-1) ) \nonumber \\
    &- \mathbf{L}^\ell \cdot \text{LN}(\mathbf{z}^\ell(t-1)) + \mathbf{\epsilon}^\ell_r \Big) + (1 - \beta) \mathbf{z}^\ell(t-1) \mbox{.} 
\end{align}
The lateral synaptic weight matrix $\mathbf{L}^\ell$ was also noted to be further decomposed in the following manner:
\begin{align}
    \mathbf{L}^\ell = \mathbf{\hat{L}}^\ell \odot \mathbf{M}^\ell \odot (1 - \mathbf{I}) + \mathbf{\hat{L}}^\ell \odot \mathbf{I}
\end{align}
where $\mathbf{I}^\ell \in \{0,1\}^{J_\ell \times J_\ell}$ is an identity matrix. 
The second term $\mathbf{\hat{L}}^\ell \odot \mathbf{I}$ creates a self-excitation effect, i.e., winning neurons suppress the activities of the losing neurons within the group, while the first term $\mathbf{\hat{L}}^\ell \odot \mathbf{M}^\ell \odot (1 - \mathbf{I})$ creates a cross-inhibition effect (details on the design of matrix $\mathbf{M}^\ell$ provided later). Note that $\mathbf{\hat{L}}^\ell$ is enforced to be strictly positive, which means that after random initialization as well as after every time it is updated, we enforce this by applying the linear rectifier function as $\mathbf{\hat{L}}^\ell(t+1) = \text{ReLU}(\mathbf{\hat{L}}^\ell(t))$. 
We commented that $\mathbf{\hat{L}}^\ell \in \mathcal{R}^{J_\ell \times J_\ell}$ is a learnable parameter matrix and $\mathbf{M}^\ell \in \{0,1\}^{J_\ell \times J_\ell}$ is a binary masking matrix that was meant to enforce a particular lateral neural competition pattern. 

The masking matrix $\mathbf{M}^\ell$ provides a valuable and flexible way for the designer/experimenter/modeler to craft different types of competition schemes across neural units within any layer of PFF's representation circuit. While there are countless ways in which this matrix could be programmed (ranging from simple winner-take-all schemes to complex neural group patterns similar to those designed for neurobiological computational models), we opted to design a ``neural-column'' form of competition (cross-layer inhibition), i.e., a scheme where neurons within pre-determined groups would compete with each (a layer contains  multiple sub-groups) resulting in an emergent, synaptic-driven form of $K$ winners-take-all. 
Formally, $\mathbf{M}^\ell$ is created according to the formula: $\mathbf{M}^\ell = \mathbf{\widehat{M}}^\ell \odot (1 - \mathbf{I})$. $\mathbf{\widehat{M}}^\ell \in \{0,1\}^{J_\ell \times J_\ell}$ is set by the experimenter (placing ones in slots where it is desired for neuronal pairs to laterally inhibit one another) according to the following process: 
\begin{enumerate}[noitemsep,nolistsep]
\item create $J_\ell / K$ matrices of shape $J_\ell \times K$ of zeros, i.e., $\{\mathbf{S}_1, \mathbf{S}_2, \cdots, \mathbf{S}_k, \cdots, \mathbf{S}_{C}\}$ (where $C = J_\ell / K$)
\item in each matrix $\mathbf{S}_k$  insert ones at all combinations of coordinates $c = \{ 1, \cdots, k, \cdots, K \}$ (column index) and $r = \{ 1 + K * (k-1), \cdots, k + K * (k-1), \cdots, K + K * (k-1) \}$ (row index)
\item concatenate the $J_\ell / K$ matrices along the horizontal axis, i.e., $\mathbf{\widehat{M}}^\ell = <\mathbf{S}_1, \mathbf{S}_2, \cdots, \mathbf{S}_C>$.
\end{enumerate}

\begin{algorithm*}[!t]
\caption{The predictive forward-forward (PFF) credit assignment algorithm. \textcolor{red}{red} denotes representation circuit computation and \textcolor{blue}{blue} denotes generative circuit computation. }
\label{alg:inference}
\fontsize{8.5}{9}\selectfont
\begin{algorithmic}[1]
   \State {\bfseries Input:} sample $(\mathbf{y}_i,\mathbf{x}_i)$, data label $c_i$ (binary label: $1 =$ ``positive'', $0 =$ ``negative''), PFF parameters $\Theta_r$ and $\Theta_g$
   \State {\bfseries Hyperparameters:} State interpolation $\beta$, SGD step size $\eta$, noise scales $\sigma_r$ and $\sigma_z$, stimulus time $T$
   \LineComment Note that $\text{LN}(\mathbf{z}^\ell) = \mathbf{z}^\ell/(||\mathbf{z}^\ell||_2 + 1\mathrm{e}{-8})$ and $\leftarrow$ denotes the overriding of a variable/object
   \Function{Simulate}{$(\mathbf{y}_i,\mathbf{x}_i, c_i), \Theta_r, \Theta_g$}
   		\LineComment Run forward pass to get initial activities
   		\State $\mathbf{z}^0 = \mathbf{x}_i$, \; $\mathbf{z}^\ell = \phi^\ell(\mathbf{W}^\ell \cdot \mathbf{z}^{\ell-1})$, for $\ell = 1,2,...,L$, \; $\mathbf{z}^{L+1} = \mathbf{y}_i$, \; $\mathbf{z}_s = \mathbf{\widehat{z}}^{L+1} = \mathbf{0}$ 
     
   		\For{$t = 1$ to $T$}
     \color{red}
   		\LineComment Run representation circuit
   		\For{$\ell = 1$ to $L$} \Comment{Compute representation activities with layer-wise parameters $\Theta^\ell_r = \{\mathbf{W}^\ell,\mathbf{V}^\ell,\mathbf{\hat{L}^\ell}\}$}
            \State $\epsilon^\ell_r \sim \mathcal{N}(0,\sigma_r)$
            \State $\mathbf{z}^\ell(t) =  \beta \Big( \phi^\ell \big( \mathbf{W}^\ell \cdot \text{LN}( \mathbf{z}^{\ell-1}(t-1) ) + \mathbf{V}^\ell \cdot \text{LN}( \mathbf{z}^{\ell+1}(t-1) ) \big) - \mathbf{L}^\ell \cdot \text{LN}(\mathbf{z}^\ell(t-1)) + \mathbf{\epsilon}^\ell_r \Big) + (1 - \beta) \mathbf{z}^\ell(t-1)$
            \State Calculate local goodness loss $\mathcal{L}(\Theta^\ell_r)$ (Equations 1 
            or 2 
            using data label $c_i$)
            \State $\Delta \mathbf{W}^\ell = \Big( 2 \frac{\partial \mathcal{L}(\Theta^\ell_r)}{\partial \sum^{J_\ell}_j (\mathbf{z}^\ell_j)^2} \odot \mathbf{z}^\ell \Big) \cdot \big( \text{LN}(\mathbf{z}^{\ell-1}) \big)^T$
            \State $\Delta \mathbf{V}^\ell = \Big( 2 \frac{\partial \mathcal{L}(\Theta^\ell_r)}{\partial \sum^{J_\ell}_j (\mathbf{z}^\ell_j)^2} \odot \mathbf{z}^\ell \Big) \cdot \big( \text{LN}(\mathbf{z}^{\ell+1}) \big)^T$
            \State $\Delta \mathbf{\hat{L}}^\ell = \Big( \Big( 2 \frac{\partial \mathcal{L}(\Theta^\ell_r)}{\partial \sum^{J_\ell}_j (\mathbf{z}^\ell_j)^2} \odot \mathbf{z}^\ell \Big) \cdot \big( \text{LN}(\mathbf{z}^{\ell}) \big)^T \Big) \odot \frac{\partial \mathbf{L}^\ell}{\partial \mathbf{\hat{L}}^\ell}$
            \State $\mathbf{W}^\ell \leftarrow \mathbf{W}^\ell - \eta \Delta \mathbf{W}^\ell$, 
            \; 
            $\mathbf{V}^\ell \leftarrow \mathbf{V}^\ell - \eta \Delta \mathbf{V}^\ell$  
            \; 
            $\mathbf{\hat{L}}^\ell \leftarrow \mathbf{\hat{L}}^\ell - \eta \Delta \mathbf{\hat{L}}^\ell$
            \Comment SGD w/ step size $\eta$ (or use Adam \cite{kingma2014adam})
    	\EndFor
     \color{blue}
    	\LineComment Run generative circuit
            \For{$\ell = L$ to $0$} \Comment Compute generative predictions with layer-wise parameters $\Theta^\ell_g = \{\mathbf{G}^{\ell+1}\}$
                \If{$\ell < L$}
                \State $\epsilon^\ell \sim \mathcal{N}(0,\sigma_z)$, \;  $\mathbf{\widehat{z}}^{\ell+1} = \phi^{\ell+1}(\mathbf{z}^{\ell+1} + \epsilon^{\ell+1})$
                \EndIf
                \State $\mathbf{\bar{z}}^\ell = \phi^\ell(\mathbf{G}^{\ell+1} \cdot \text{LN}(\mathbf{\widehat{z}}^{\ell+1}))$
                \State Calculate local generative loss $\mathcal{L}^\ell_g(\mathbf{G}^{\ell+1}) = \frac{1}{2} \sum_j (\mathbf{\bar{z}}^\ell_j - \mathbf{z}^\ell_j(t))^2$
                \State $\mathbf{e}^\ell = \mathbf{\bar{z}}^\ell - \mathbf{z}^\ell$, \; $\Delta \mathbf{G}^{\ell+1} = \mathbf{e}^\ell \cdot \big( \text{LN}(\mathbf{\widehat{z}}^{\ell+1}(t)) \big)^T$  \Comment{Notice that $\mathbf{e}^\ell = \frac{\partial \mathcal{L}^\ell_g(\mathbf{G}^{\ell+1})}{\partial \mathbf{\bar{z}}^\ell}$}
                \State $\mathbf{G}^{\ell+1} \leftarrow \mathbf{G}^{\ell+1} - \eta \Delta \mathbf{G}^{\ell+1}$
            \EndFor 
            \State $\mathbf{\widehat{z}}^{L+1} \leftarrow \mathbf{\widehat{z}}^{L+1} - \gamma \frac{\partial \mathcal{L}^L_g(\mathbf{G}^{L+1})}{\partial \mathbf{\widehat{z}}^{L+1}}$ 
            \Comment Update latent variable $\mathbf{z}_s$ (one step of iterative inference)
    	\EndFor
        \color{black}
        \State \textbf{Return} $\Theta_g = \{\mathbf{G}^1,...,\mathbf{G}^L,\mathbf{G}^{\ell+1}\}, \Theta_r = \{\mathbf{W}^1,...,\mathbf{W}^{L-1}, \mathbf{W}^L\}$ \Comment Output newly updated PFF parameters
    \EndFunction
\end{algorithmic}
\end{algorithm*}





\subsection*{Visualized Samples (Expanded)}

\begin{figure*}[!htb]
     \centering
     \begin{subfigure}[b]{0.35\textwidth}
         \centering
         \includegraphics[width=\textwidth]{figs/mnist_recon.png}
         \caption{PFF reconstructed images.}
         \label{fig:mnist_recon_big}
     \end{subfigure}
     \hspace{0.5cm}
     \begin{subfigure}[b]{0.35\textwidth}
         \centering
         \includegraphics[width=\textwidth]{figs/mnist_samples.jpg}
         \caption{PFF sampled images.}
         \label{fig:mnist_fantasies_big}
     \end{subfigure} \\
     \begin{subfigure}[b]{0.4\textwidth}
         \centering
         \includegraphics[width=\textwidth]{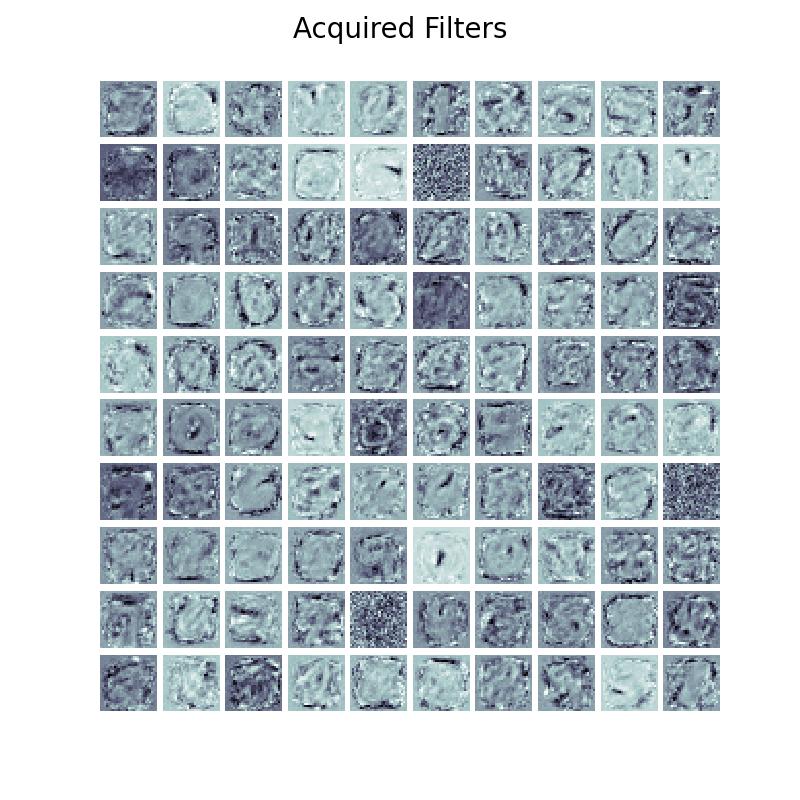}
         \caption{PFF representation receptive fields.}
         \label{fig:mnist_rep_filters_big}
     \end{subfigure}
     \begin{subfigure}[b]{0.4\textwidth}
         \centering
         \includegraphics[width=\textwidth]{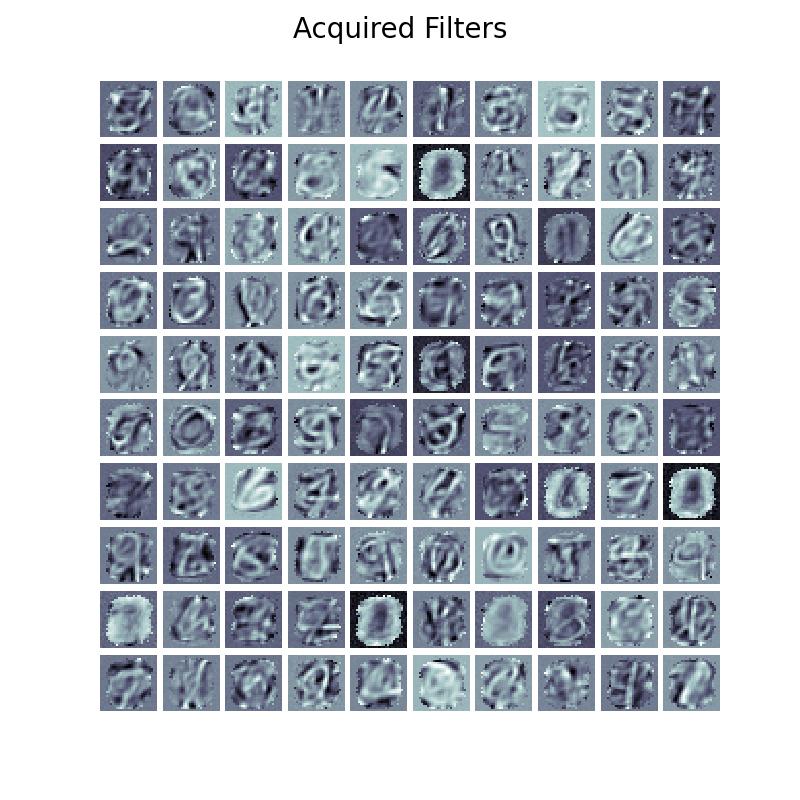}
         \caption{PFF generative receptive fields.}
         \label{fig:mnist_gen_filters_big}
    \end{subfigure}
        \caption{MNIST model reconstruction (Left) and generated (Right) samples. In the bottom row, the receptive fields of the bottom-most layer of the representation circuit (Left) and those of the generative circuit (Right).}
        \label{fig:mnist_samples_big}
\end{figure*}

\begin{figure*}[!htb] 
     \centering
     \begin{subfigure}[b]{0.35\textwidth}
         \centering
         \includegraphics[width=\textwidth]{figs/kmnist_recon.png}
         \caption{PFF reconstructed images.}
         \label{fig:kmnist_recon_big}
     \end{subfigure}
     \hspace{0.5cm}
     \begin{subfigure}[b]{0.35\textwidth}
         \centering
         \includegraphics[width=\textwidth]{figs/kmnist_samples.jpg}
         \caption{PFF sampled images.}
         \label{fig:kmnist_fantasies_big}
     \end{subfigure} \\
     \begin{subfigure}[b]{0.4\textwidth}
         \centering
         \includegraphics[width=\textwidth]{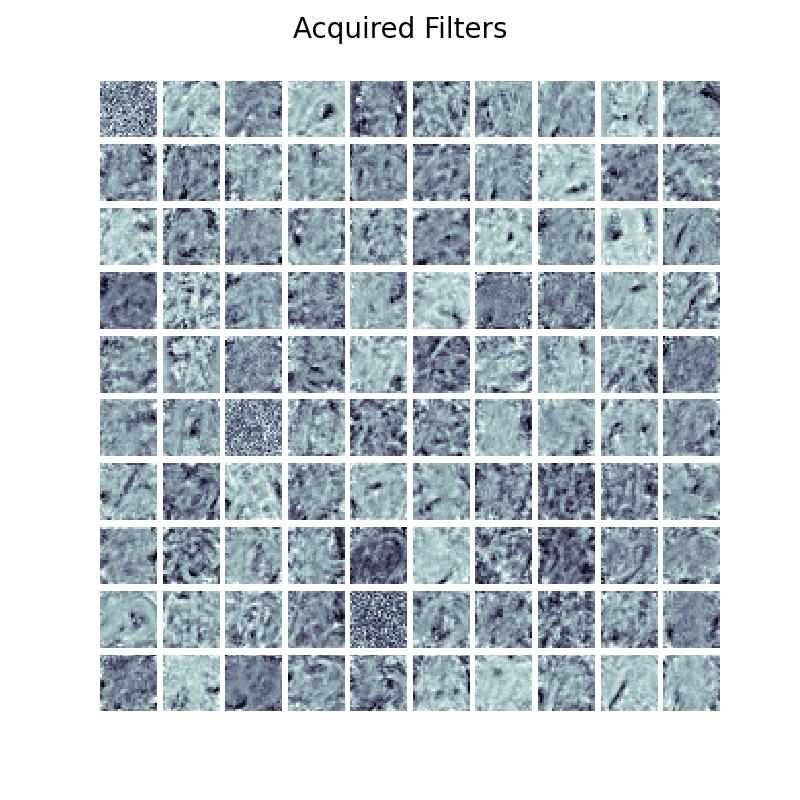}
         \caption{PFF representation receptive fields.}
         \label{fig:kmnist_rep_filters_big}
     \end{subfigure}
     \begin{subfigure}[b]{0.4\textwidth}
         \centering
         \includegraphics[width=\textwidth]{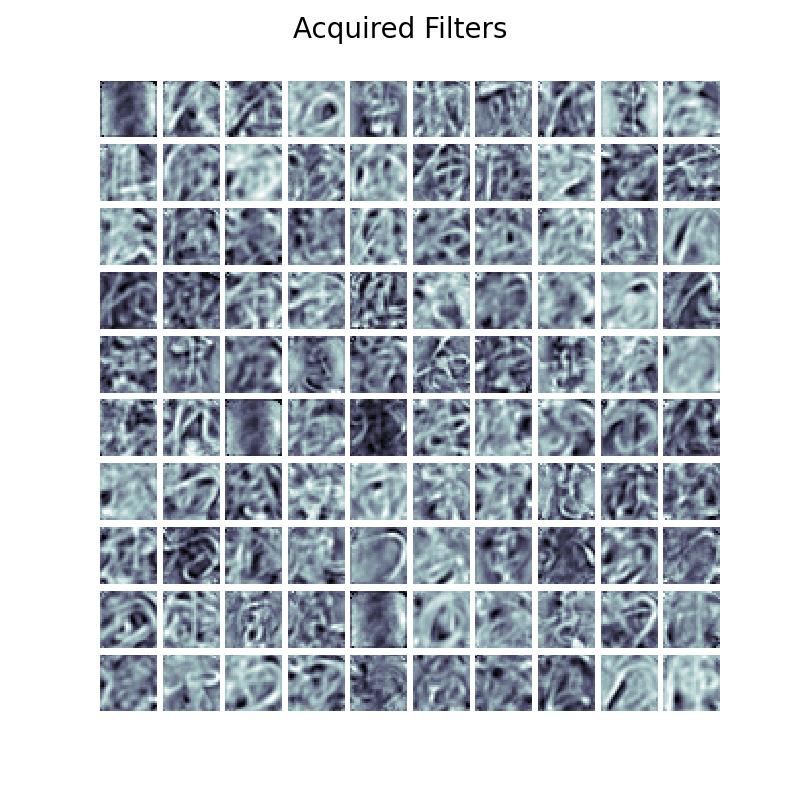}
         \caption{PFF generative receptive fields.}
         \label{fig:kmnist_gen_filters_big}
     \end{subfigure}
        \caption{In the top row, Kuzushiji-MNIST model reconstruction (Left) and generated (Right) samples. In the bottom row, the receptive fields of the bottom-most layer of the representation circuit (Left) and those of the generative circuit (Right).}
        \label{fig:kmnist_samples_big}
\end{figure*}

Figures \ref{fig:mnist_samples_big} and \ref{fig:kmnist_samples_big} present the reconstruction and synthesized samples from the PFF models in the main paper at a larger image scale/size for both MNIST and K-MNIST. 
Furthermore, in the bottom rows of each figure, we visualize the receptive fields acquired by PFF for each of these two datasets. Observe that the receptive fields (of the synapses of the layer closest to the sensory input layer) acquired by  both the representation and generative circuits appear to extract useful/interesting structure related to a digit or Kanji character strokes, often, as is expected for fully-connected neural structures, acquiring representative full object templates (if one desired each receptive field to acquire only single strokes/component features specifically, then an additional prior would need to be imposed, such as convolution or the locally-connected receptive field structure employed in \cite{bartunov2018assessing,hinton2022forward}).

\subsection*{Dataset Details}
In the main paper, we experimented with several (gray-scale) image collections, i.e., the MNIST, Kuzushiji-MNIST (K-MNIST), Fashion MNIST (F-MNIST), Not-MNIST (N-MNIST), and Ethiopic (Et-MNIST) databases. All of these databases contained $28\times28$ images from $10$ different categories. 
The MNIST dataset \cite{lecun1998mnist} specifically contains images containing handwritten digits across $10$ different categories. Kuzushiji-MNIST is a challenging drop-in replacement for MNIST, containing images depicting hand-drawn Japanese Kanji characters \cite{clanuwat2018deep} (each class corresponding to the character's modern hiragana counterpart, with $10$ classes in total). 
Fashion MNIST \cite{xiao2017fashion} is a challenging drop-in replacement for MNIST, contains image patterns depicting clothing items out of $10$ item classes. 
The NotMNIST database\footnote{http://yaroslavvb.blogspot.com/2011/09/notmnist-dataset.html} is a more difficult variation of MNIST created by replacing the digits with characters of varying fonts/glyphs (letters A-J). 
Finally, Ethiopic MNIST is a database that contains ten distinct numerals (generated via elastic deformations of Unicode numeral exemplars) for the Ge`ez language (specifically, this database contains patterns for the numerals one through ten, as no numeral for zero exists), one of four orthographies contained in the low-resource glyph/numeral recognition dataset proposed in \cite{wu2020afromnist}.

\end{document}